\documentclass[journal]{IEEEtran}
\usepackage{amsmath, amsfonts, amssymb, bm} 
\usepackage{newtxmath} 
\usepackage{enumitem}
\usepackage{algorithmic}
\usepackage{array}
\usepackage[caption=false,font=footnotesize]{subfig}
\usepackage{textcomp}
\usepackage{stfloats}
\usepackage{url}
\usepackage{booktabs}
\usepackage{verbatim}
\usepackage{graphicx}
\usepackage{epstopdf} 
\def\BibTeX{{\rm B\kern-.05em{\sc i\kern-.025em b}\kern-.08em
    T\kern-.1667em\lower.7ex\hbox{E}\kern-.125emX}}
\usepackage{balance}
\usepackage[commandnameprefix=ifneeded]{changes}
\usepackage{soul}

\begin{document}
\title{A Flow-Theoretic Multimodal Control Framework Fusing Mental and Physical Engagement for Upper-Limb Rehabilitation}
\author{
Jun Leng, \IEEEmembership{Student Member, IEEE}, Pengcheng Li, \IEEEmembership{Senior Member, IEEE}, Hanze Wang, \IEEEmembership{Student Member, IEEE}, Shuxiang Guo, \IEEEmembership{Fellow, IEEE}, and Junlin Jiang, \IEEEmembership{Student Member, IEEE}
\thanks{This work has been submitted to the IEEE for possible publication. Copyright may be transferred without notice, after which this version may no longer be accessible.}
\thanks{Manuscript received [ ]. This work was supported by the High Level of Special Funds from the Southern University of Science and Technology under Grant G03034K003, and in part by the Shenzhen ``Medical Engineering Integration'' Special Support Program Project under Grant F-2024-Z99-502365. \textit{(Corresponding authors: Pengcheng Li; Shuxiang Guo.)}}
\thanks{Jun Leng, Pengcheng Li, Hanze Wang, Shuxiang Guo, and Junlin Jiang are with the Department of Electronic and Electrical Engineering, Southern University of Science and Technology, Shenzhen 518055, China (e-mail: lengj2024@mail.sustech.edu.cn, lipc@sustech.edu.cn; guo.shuxiang@sustech.edu.cn).}
\thanks{Shuxiang Guo is also with the Aerospace Center Hospital, School of Life Science, and the Key Laboratory of Convergence Medical Engineering System and Healthcare Technology, Ministry of Industry and Information Technology, Beijing Institute of Technology, Beijing 100081, China.}
}

\markboth{IEEE Transactions on Human-Machine Systems (Under Review)}%
{Leng \MakeLowercase{\textit{et al.}}: A Flow-Theoretic Multimodal Control Framework for Upper-Limb Rehabilitation}

\maketitle

\begin{abstract}
While Assist-as-Needed (AAN) control aims to maximize user participation in robot-assisted training, existing methods relying solely on physical metrics cannot distinguish muscular inability from mental disengagement, leading to suboptimal interventions. To address this ambiguity, a Flow-theoretic multimodal AAN framework is proposed, fusing EEG-based mental engagement and sEMG-based physical engagement into an adaptive impedance control strategy. A four-quadrant fuzzy decision engine translates these dual physiological metrics into continuous control policies. This enables the upper-limb exoskeleton to inject feedforward assistance during intent-execution mismatch (e.g., motor output attenuation despite mental focus) and escalate guiding stiffness to penalize mental distraction. The framework was validated with healthy participants against a state-of-practice sEMG-only AAN baseline. Kinematically and mechanically, the proposed framework reduced robotic energy expenditure by 71.9\% while maintaining precise trajectory tracking, confirming it elicited greater voluntary human effort. Physiologically, the system increased the time spent in the neuroplasticity zone (Flow State Ratio) from 4.6\% to 42.8\%. These results validate that decoupling and jointly regulating mental and physical states enhances the selectivity of robotic assistance, mitigating user passivity and sustaining the engagement crucial for motor learning.
\end{abstract}


\section{Introduction}
\label{sec:introduction}

\IEEEPARstart{U}{pper} limb movement dysfunction caused by neurological injuries, such as stroke, severely impairs an individual’s ability to perform Activities of Daily Living (ADLs) and significantly reduces quality of life \cite{science1}. Modern neurorehabilitation relies on promoting patients’ active participation in high-intensity, repetitive training to stimulate neuroplasticity and facilitate functional recovery \cite{zhudong1}. To address these rigorous demands, robot-assisted therapy provides consistent, quantifiable training that directly complements traditional therapist-led methods \cite{lpc,whz}. These systems offer precise physical guidance while reducing the labor burden on clinicians, making them essential for delivering the high-dosage therapy required for motor learning.

To maximize therapeutic outcomes, the field has widely adopted the ``Assist-as-Needed'' (AAN) paradigm. The core principle of AAN is to minimize robotic intervention, providing assistance only when the user cannot complete the movement independently. This approach aims to ensure maximal active effort \cite{ann1}. However, accurately inferring the user's true ``need'' remains a critical challenge. Traditional AAN strategies rely heavily on physical performance metrics, such as kinematic error or interaction force. For instance, Han et al. \cite{ann2} proposed an adaptive control method that infers movement intention by decomposing real-time interaction forces. The system then adjusts the robot's impedance parameters to promote active participation. More recently, researchers have integrated surface Electromyography (sEMG) to incorporate physiological effort into the control loop. Xu et al. \cite{ann3}, for example, developed a framework that adapts training trajectories and robotic assistance based on sEMG-derived muscle activation. This system encourages patient participation by modulating feedback gains. However, while such approaches successfully measure mechanical or muscular execution, they still fail to capture user's mental focus. Although high-amplitude sEMG quantifies physical exertion, it cannot differentiate genuine task engagement from pathological co-contraction or abnormal compensatory synergies. Execution-level metrics alone fall short in revealing the user's mental intent. Without monitoring mental focus, purely physical assistance strategies often fail to maintain an optimal training challenge, eventually leading to user passivity.

To resolve this ambiguity, integrating electroencephalography (EEG) into the control loop is essential for providing a direct measure of task-related attentional focus \cite{eeg}. While EEG is widely applied in robotic rehabilitation, current paradigms mostly use it within Motor Imagery-based Brain-Computer Interfaces (MI-BCI) to decode discrete movement intentions \cite{mibci1}, \cite{mibci2}. 

This paper introduces an adaptive control framework grounded in Flow Theory to map these multidimensional physiological signals into actionable robotic interventions. Classic Flow Theory defines an optimal mental state as a balance between external task challenge and intrinsic user skill \cite{flow1, flow2}. The modern four-quadrant extension categorizes user experience into Flow, Anxiety, Boredom, and Apathy based on these misalignments \cite{flow_four}. In robot-assisted training, keeping users within this flow state mitigates both task-induced anxiety and under-stimulated boredom \cite{eeg4flow1, eeg4flow2}. This equilibrium sustains intrinsic motivation and facilitates motor learning \cite{flow2_plasticity}. 

Most existing flow-aware systems rely on predefined task difficulties or intermittent self-reporting, limiting their real-time adaptability. Directly quantifying the abstract Challenge and Skill axes in real-time is also highly impractical during dynamic robot-assisted training, as the user's motor capabilities fluctuate continuously due to fatigue and varying physical demands. To address this gap, the proposed framework bypasses the direct measurement of classic Flow coordinates. It instead establishes a two-dimensional physiological diagnostic space using mental engagement ($s^*_{\text{me}}$, via EEG) and physical engagement ($s^*_{\text{pe}}$, via sEMG) as continuous proxies. By continuously monitoring the congruence or mismatch between cognitive intent and motor execution, the system inversely infers the user's psychological state. This diagnostic mapping directly drives the real-time, closed-loop interventions. The main contributions of this paper are summarized as follows:

\begin{enumerate}[label=\arabic*), leftmargin=*]
    \item A multimodal, mental-physical aware exoskeleton system is developed to fuse real-time EEG and sEMG signals, enabling the continuous monitoring of the user's mental and physical engagement during robot-assisted training.
    
    \item A Flow-grounded fuzzy decision engine, coupled with a sensorless variable impedance controller, is proposed. By estimating the user's Flow state, the system dynamically adjusts guiding stiffness and feedforward assistance to mitigate physical passivity, address the mismatch between mental intent and physical execution, and optimize the AAN interaction.
\end{enumerate}

\section{System Overview and Hardware Platform}

To validate the proposed flow-grounded control framework, the Cable-driven Upper-limb Rehabilitation Exoskeleton (CURE) is utilized. This platform, developed in previous work \cite{pingtai}, is integrated with a multimodal physiological sensing suite.

\subsection{CURE Exoskeleton}
\begin{figure}[htbp]
    \centering
    
    \captionsetup[subfloat]{captionskip=2pt}

    \subfloat[]{\includegraphics[width=0.42\linewidth]{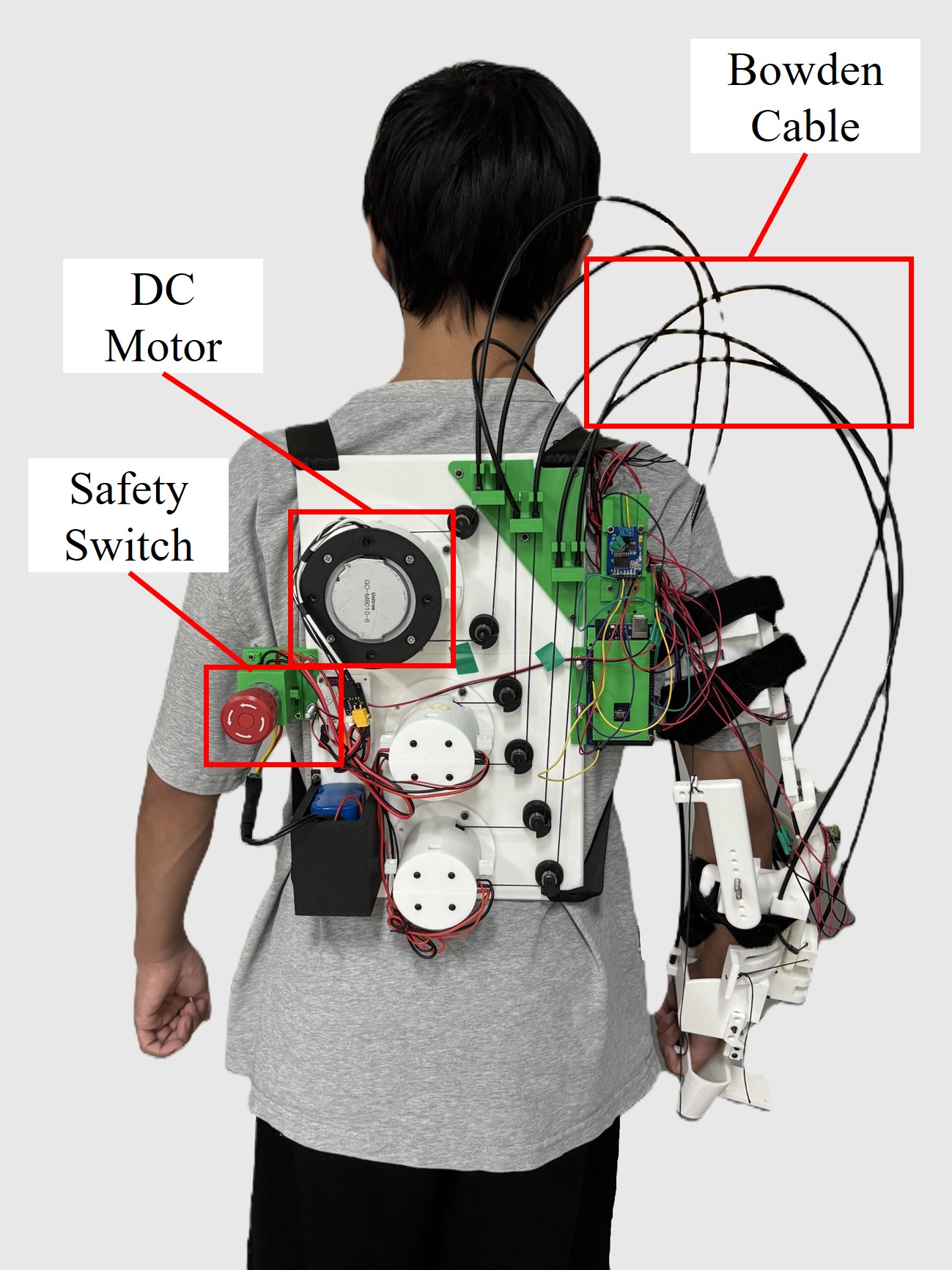}\label{fig:sub1}}
    \hspace{0.04\linewidth} 
    \subfloat[]{\includegraphics[width=0.42\linewidth]{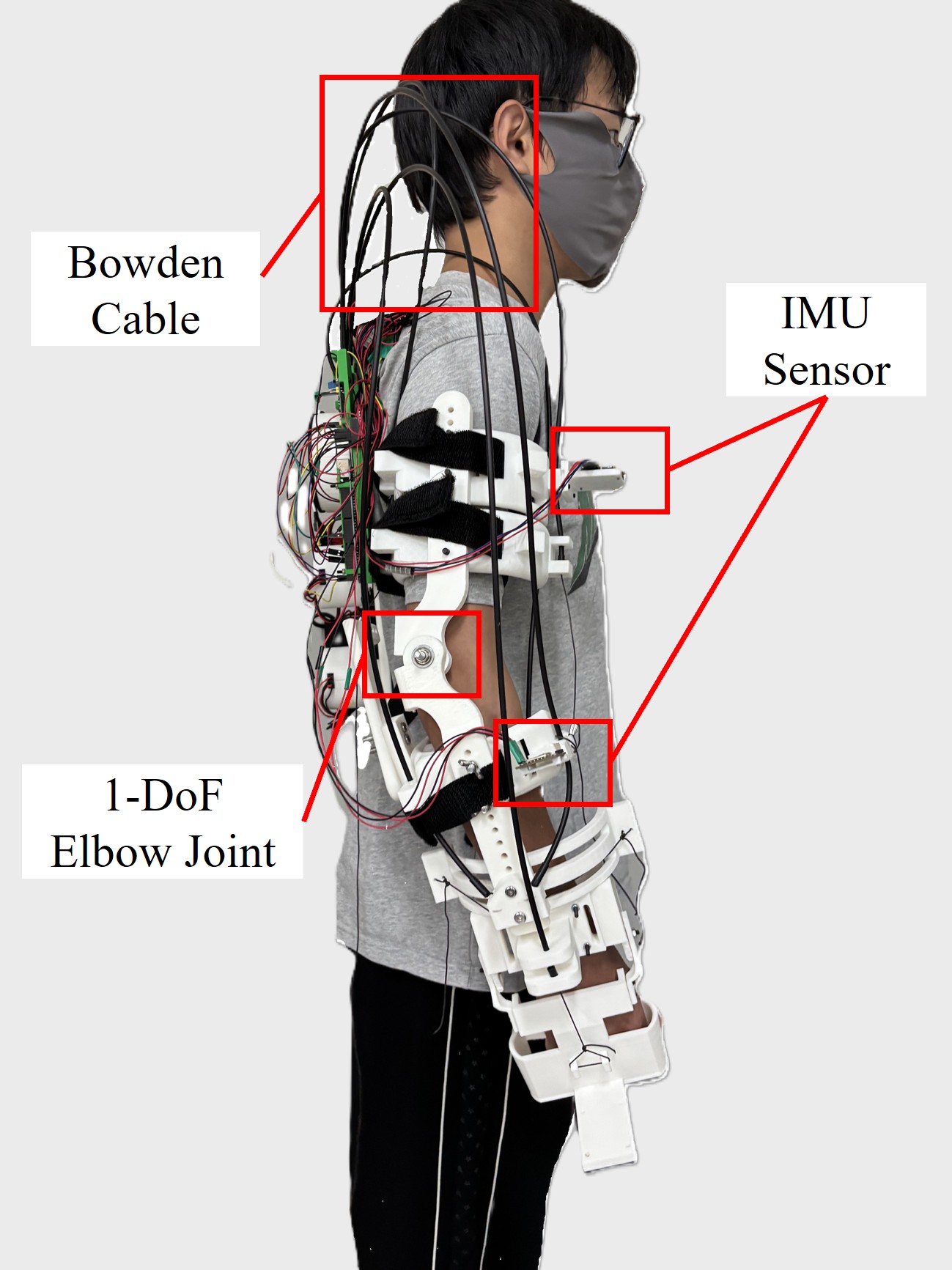}\label{fig:sub2}} \\ 
    
    \vspace{0.2cm} 
    
    \subfloat[]{\includegraphics[width=0.84\linewidth]{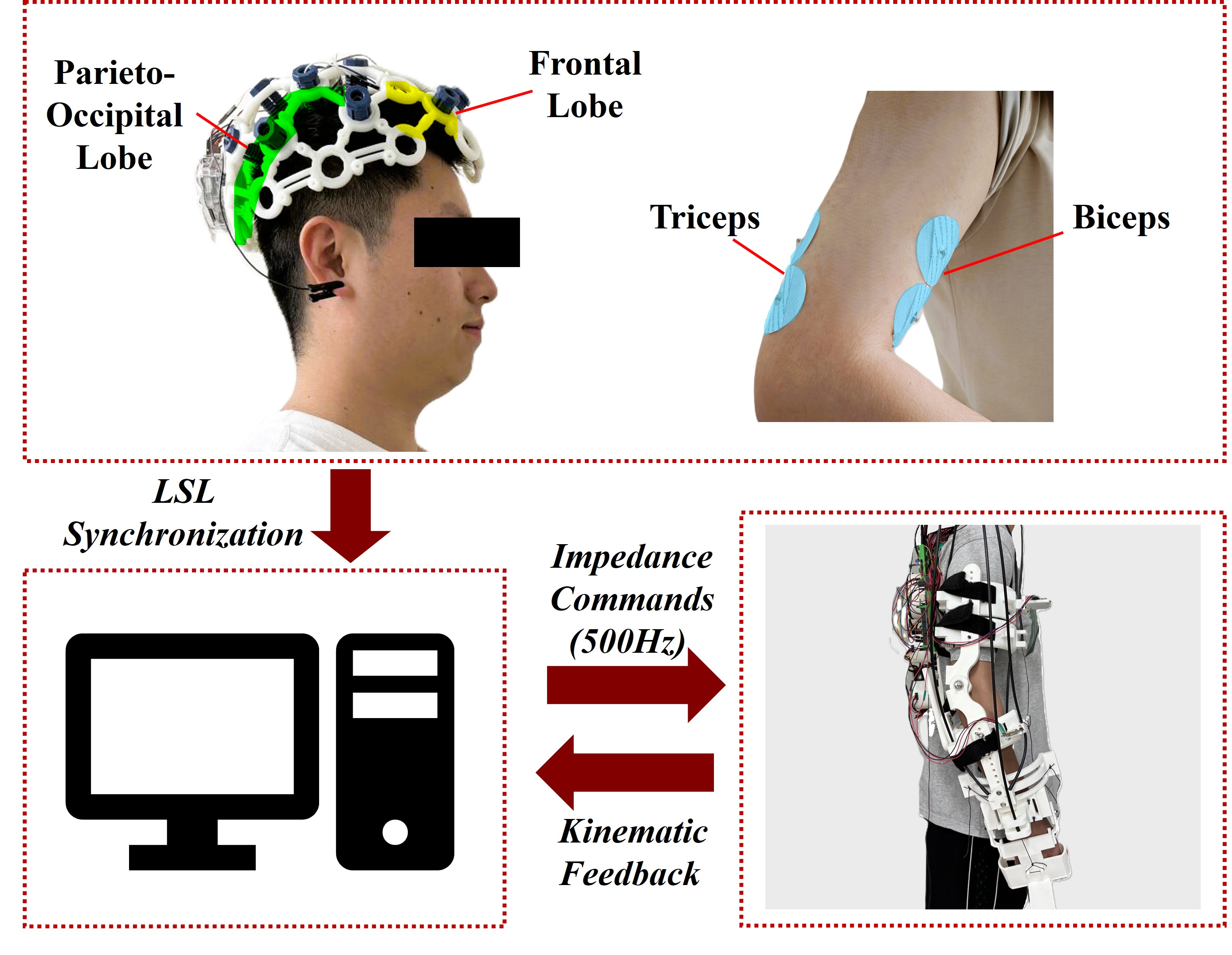}\label{fig:sub3}}

    \caption{System overview and hardware architecture. (a) Posterior view showing the backpack-mounted actuation unit and the Bowden-cable transmission. (b) Lateral view of the user wearing the CURE exoskeleton, detailing the 1-DoF elbow joint and IMU sensor placements. (c) Schematic of the multimodal sensing interface (EEG and sEMG) and the closed-loop data flow architecture.}
    \label{fig:yjian}
\end{figure}

The CURE system targets home-based rehabilitation (Fig.~\ref{fig:yjian}(a), (b)). A Bowden-cable transmission spatially separates the actuation unit from the wearable arm brace, relocating the heavy components to a backpack. This configuration reduces the distal arm mass to approximately 0.7\,kg, which lowers interaction inertia and delays user fatigue. Although the robot supports three degrees of freedom (DoFs) across the elbow and wrist, this study validates the control strategy on the 1-DoF elbow joint. 

A high-torque DC motor (GO-M8010-6, Unitree, China) actuates this joint. Motor encoders and Inertial Measurement Units (IMUs, JY901, WitMotion, China) measure joint kinematics (angular position $q$ and velocity $\dot{q}$) to provide high-frequency state feedback for the closed-loop controller. A backpack-mounted emergency stop button provides a hardware safety override.

\subsection{Multimodal Sensing Interface}
A multimodal sensing interface (Fig.~\ref{fig:yjian}(c)) is integrated to close the human-in-the-loop control. A wireless 8-channel EEG system (OpenBCI Cyton, USA) captures real-time neural activity, with dry electrodes positioned at Fp1, Fp2, F3, F4, Fz, Cz, Pz, and Oz. This montage targets frontal and parieto-occipital regions to decode mental engagement. Bipolar sEMG sensors (Sichiray, China) attached to the Biceps and Triceps Brachii quantify physical engagement by monitoring agonist voluntary effort and antagonist co-activation. A central PC manages the entire system, acquiring physiological streams via LSL and serial interfaces and synchronizing them with the robot's low-level torque controller at a 100\,Hz loop rate.

\begin{figure*}[t]
    \centering
    \includegraphics[width=\textwidth]{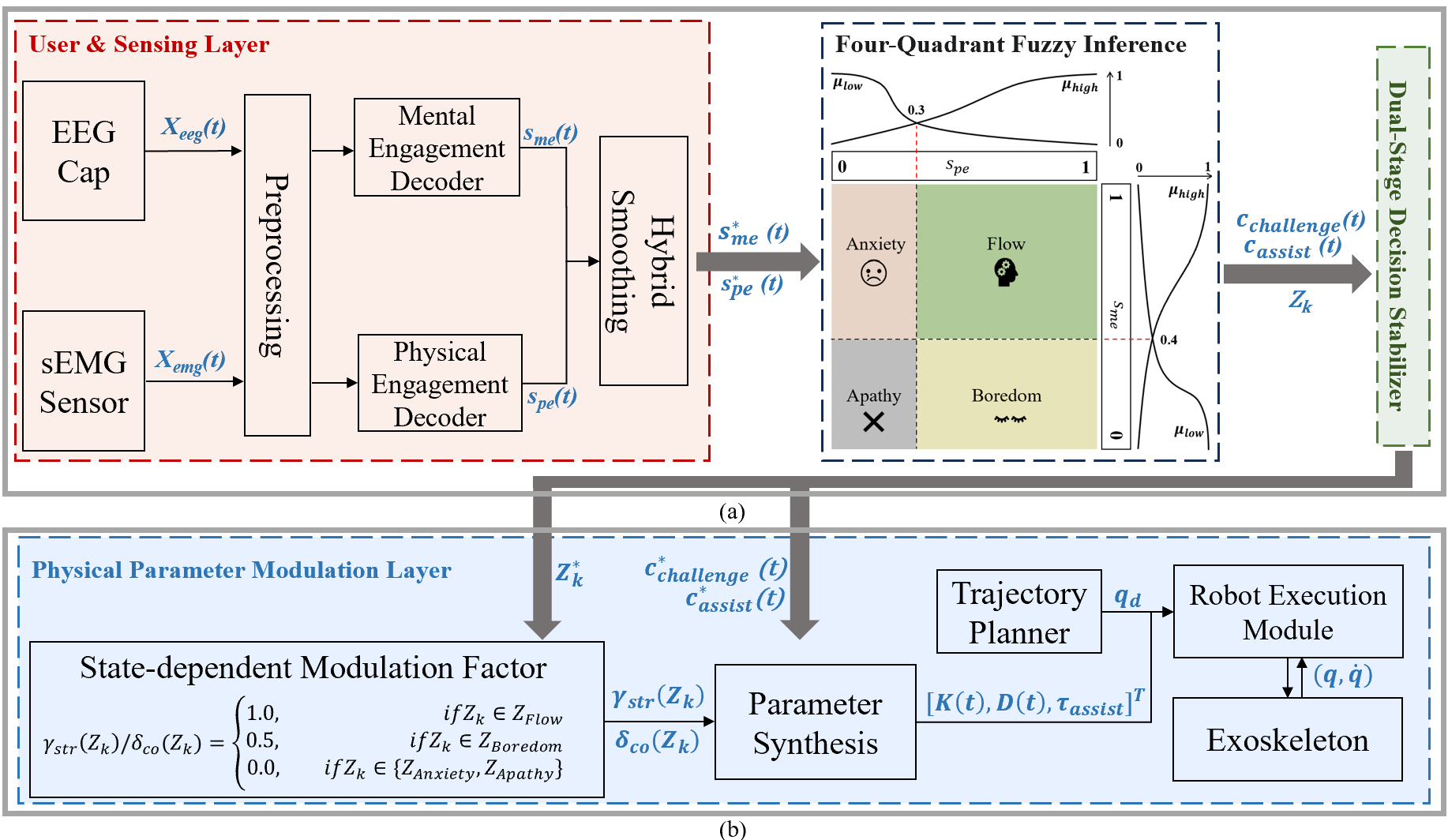}
    \caption{Overall architecture of the proposed flow-grounded AAN framework. (a) High-Level Controller: A multimodal decision-making engine that processes EEG and sEMG signals to map the user's mental and physical engagement into continuous strategic policies via fuzzy inference. (b) Low-Level Controller: An adaptive execution module that synthesizes these policies into targeted impedance parameters ($K, D, \tau_{\text{assist}}$) to regulate sensorless human-robot interaction.}
    \label{fig:quadrant_model}
\end{figure*}

\section{Flow-Grounded Strategy and Decision Making}
The proposed hierarchical framework (Fig.~\ref{fig:quadrant_model}) is decoupled into two primary components. Grounded in Flow Theory, the high-level controller (Fig.~\ref{fig:quadrant_model}(a)) acts as the multimodal decision-making engine, evaluating physiological states to balance task challenge with user skill. The low-level controller (Fig.~\ref{fig:quadrant_model}(b)) translates these strategic policies into adaptive impedance parameters for physical execution.

\subsection{Multimodal Physiological Decoding} 
To quantify real-time internal states, a multimodal interface fusing EEG and sEMG is utilized, applying hybrid smoothing to ensure robust control inputs.

\textit{1) Signal Preprocessing:} EEG (250\,Hz) and sEMG (1000\,Hz) signals are preprocessed using 50\,Hz notch filters, alongside 1--40\,Hz FIR and 20--450\,Hz Butterworth band-pass filters, respectively. Following sEMG full-wave rectification, both streams are segmented via a 1.0\,s sliding window (0.25\,s step). Within each epoch, sEMG intensity is quantified via root-mean-square (RMS) amplitude, and EEG power spectral density (PSD) is estimated via Welch's method (0.5\,s sub-windows).

\textit{2) Mental Engagement ($s^*_{\text{me}}$):} As a proxy for mental focus \cite{attention1}, this metric integrates the normalized Theta/Beta Ratio ($\hat{R}_{\text{TBR}}(t) \in [0, 1]$) and Alpha Suppression Index ($\hat{I}_{\alpha}(t) \in [0, 1]$) from frontal EEG channels into a weighted combination: 
\begin{equation} 
s_{\text{me, raw}}(t) = w_{\text{tbr}} \cdot \hat{R}_{\text{TBR}}(t) + w_{\alpha} \cdot \hat{I}_{\alpha}(t)
\end{equation} 
where positive weights satisfy $w_{\text{tbr}} + w_{\alpha} = 1$. To mitigate transient artifacts and attentional shifts, a moving average filter yields a stable, tonic metric $s^*_{\text{me}}(t)$ suitable for continuous robotic modulation.

\textit{3) Physical Engagement ($s^*_{\text{pe}}$) and Co-activation ($s_{\text{co}}$):} Two motor metrics are derived from sEMG. The primary metric, physical engagement ($s^*_{\text{pe}}$), quantifies voluntary output via RMS amplitude normalized by Maximum Voluntary Contraction (MVC) \cite{activity_level}. Letting $\bar{s}_{\text{agonist}}(t)$ and $\bar{s}_{\text{antagonist}}(t)$ denote normalized activations, the raw metric penalizes excessive co-contraction: 
\begin{equation} 
s_{\text{raw}}(t) = \frac{1}{100} \cdot \max \left(0, \bar{s}_{\text{agonist}}(t) - w \cdot \bar{s}_{\text{antagonist}}(t)\right) 
\end{equation} 
where $w \in [0, 1]$ is a penalty weight. To prevent false disengagement detection during rhythmic relaxation, a peak-hold mechanism with exponential decay yields the smoothed score $s^*_{\text{pe}}(t)$, providing the core input for the fuzzy engine.

To inform low-level joint rigidity (Section IV-A), a secondary co-activation metric is defined as the contraction overlap: 
\begin{equation} 
s_{\text{co}}(t) = \min \left(\bar{s}_{\text{agonist}}(t), \bar{s}_{\text{antagonist}}(t) \right)
\end{equation}

\subsection{Flow-Grounded Theoretical Mapping}
Because quantifying absolute ``Challenge'' and ``Skill'' in real-time remains impractical, the decision-making engine translates the four-quadrant Flow model into a two-dimensional physiological diagnostic space (Fig.~\ref{fig:quadrant_model}(a)). Rather than relying on intermittent self-reporting, real-time control requires a reverse-inference approach. Mental engagement ($s^*_{\text{me}}$) proxies the user's perceived cognitive demand (task Challenge), while effective physical engagement ($s^*_{\text{pe}}$) reflects execution capability (Skill). Monitoring their congruence allows the system to infer psychological divergence from multi-modal signatures.

Neuroergonomic and cognitive workload literature demonstrates that shifts in the Challenge-Skill balance elicit distinct physiological manifestations across central and peripheral pathways. For example, concurrent high cognitive demand during physical exertion induces prefrontal neuronal interference, which increases neuromotor noise and impairs joint steadiness \cite{3b_theory1}. This provides a quantifiable basis for task overload (Challenge $\gg$ Skill). Conversely, resource theory suggests highly overlearned, automated motor executions demand minimal attentional resources; prolonged exposure to such task underload (Challenge $\ll$ Skill) diminishes cognitive vigilance while preserving physical task-tracking output \cite{3b_theory1}. When both task-directed cognitive drive and volitional motor execution are deficient, the neuromuscular system exhibits concurrent hypo-arousal and dropping active torque, characterizing complete task disengagement \cite{3b_theory3}. Finally, neurophysiological evidence shows active volitional intention coupled with effective peripheral execution induces the early cortical activation and somatosensory synchronization vital for Hebbian-based neural plasticity \cite{3b_theory2}. Maintaining the task-skill equilibrium (Challenge $\approx$ Skill) thus ensures the synchronous central and peripheral activation necessary for optimal motor learning.

By diagnosing the user through this reverse-inference lens, the system maps these four intent-execution manifestations directly to targeted robotic interventions within a closed-loop framework. 
1) Anxiety (High $s^*_{\text{me}}$, Low $s^*_{\text{pe}}$): Intense cognitive effort fails to produce effective motor output, reflecting an acute intent-execution mismatch (task challenge severely exceeds motor capability) that triggers an Assist \& Guide intervention. 
2) Boredom (Low $s^*_{\text{me}}$, High $s^*_{\text{pe}}$): High motor efficacy with minimal cognitive allocation points to rote automaticity; to counter this, a Correction \& Alert strategy (e.g., escalating virtual stiffness) is executed to prompt cortical re-engagement. 
3) Flow (High $s^*_{\text{me}}$, High $s^*_{\text{pe}}$): Synchronous, high-intensity activation across both domains indicates an optimal challenge-skill match, sustaining a Maintain \& Challenge strategy to foster motor adaptation. 
4) Apathy (Low $s^*_{\text{me}}$, Low $s^*_{\text{pe}}$): Concurrent low neural drive and muscular output signify complete task disengagement, initiating a Re-engage protocol to safely pause training.

This feature-engineered diagnostic mapping circumvents the need for explicit subjective coordinate measurement. It translates the abstract psychological dimensions of Flow Theory into computable physiological parameters for real-time, closed-loop robotic modulation.

\subsection{Decision Engine Implementation}
To translate the aforementioned qualitative strategies into continuous and safe control commands, the right section of Fig. \ref{fig:quadrant_model}(a) implements two sequential modules: a fuzzy inference engine and a decision stabilizer.

\begin{figure}[htbp]
    \centering
    \captionsetup[subfloat]{captionskip=2pt} 
    
    \subfloat[\label{fig:surf_assist}]{
        \includegraphics[width=0.47\linewidth]{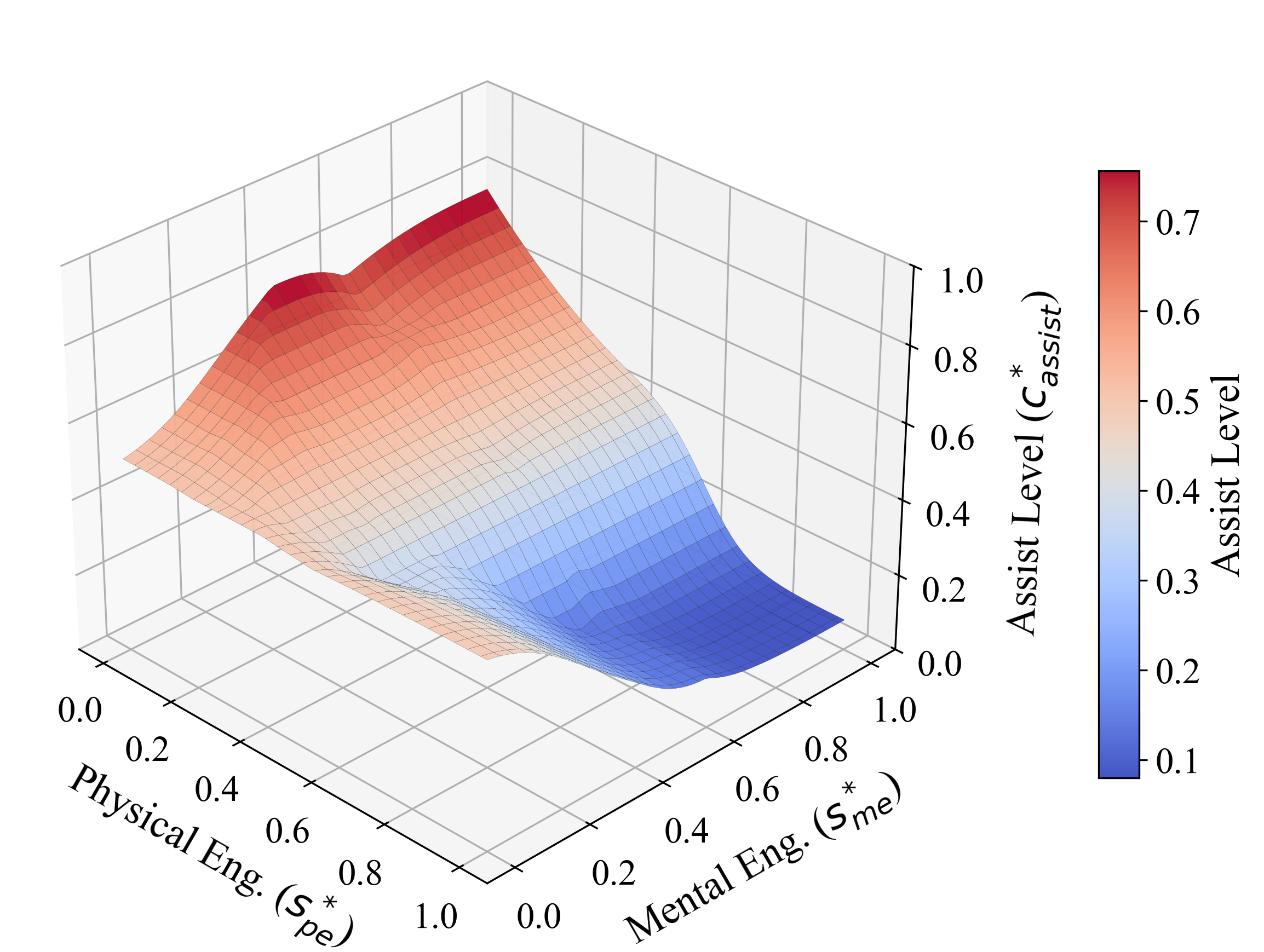}
    }
    \hfill 
    \subfloat[\label{fig:surf_challenge}]{
        \includegraphics[width=0.47\linewidth]{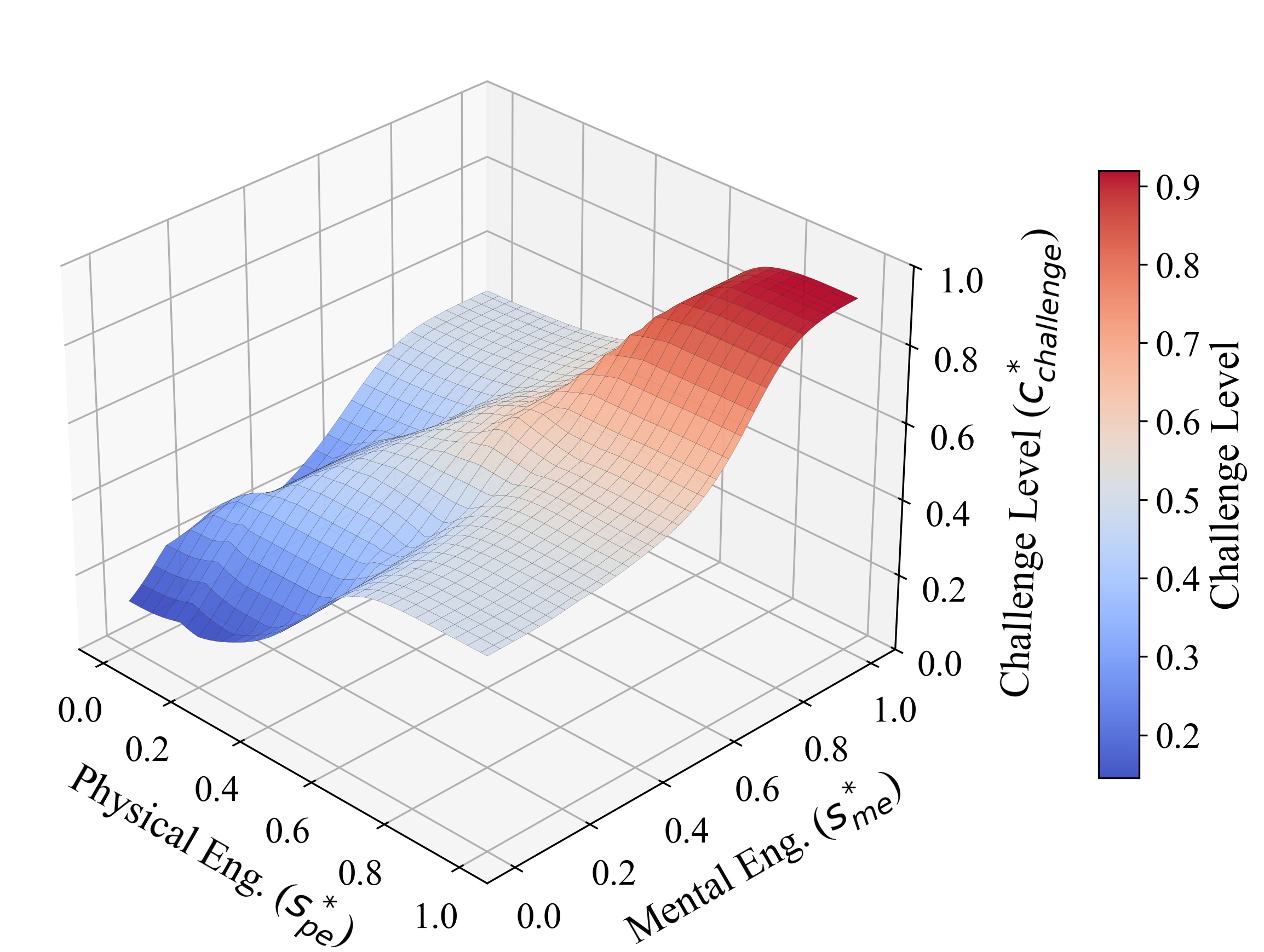}
    }
    
    \caption{Continuous control output surfaces generated by the Fuzzy Inference System. (a) The Assist Level ($c^*_{\text{assist}}$) surface mapped across the two-dimensional physiological state space. (b) The Challenge Level ($c^*_{\text{challenge}}$) surface.}
    \label{fig:fuzzy_surface_combined}
\end{figure}

\textit{1) Four-Quadrant Fuzzy Inference:} The module maps continuous physiological states ($s^*_{\text{me}}, s^*_{\text{pe}}$) to two continuous policy variables: Assistance Demand ($c^*_{\text{assist}}$) and Challenge Level ($c^*_{\text{challenge}}$). To ensure continuous differentiability and prevent mechanical chattering, Gaussian membership functions $\mu(x) = \exp(-\frac{(x - c)^2}{2\sigma^2})$ are employed. Rather than using symmetric mathematical midpoints, parameters are derived from physiological baselines. For mental engagement ($s^*_{\text{me}}$), users naturally sustain active tracking between 0.5 and 0.8, while distracted states fall below 0.3. A symmetric 0.5 threshold would be overly punitive. Thus, the Gaussian parameters ($c_{\text{med}}=0.4$, $\sigma=0.15$) position the dominant inflection point at 0.4, a boundary empirically established via subsequent preliminary tuning trials (Section V-C.1), accommodating natural attentional fluctuations while detecting drift early \cite{eeg_jiaozhun}.

A core design feature is ``Physiological Clamping'' for physical engagement ($s^*_{\text{pe}}$). Because muscle activation below 30\% MVC is insufficient to elicit robust motor learning \cite{activity_level, 30mvc}, the ``High'' state is asymmetrically parameterized ($c=1.0, \sigma=0.35$). This broad, left-skewed tail places the clamping threshold (inflection point) at 0.3, aligning with the preliminary experimental results. Consequently, the system classifies any muscular output below 0.3 as low engagement. This baseline restricts sub-optimal effort, forcing the system to either provide assistance (during Anxiety) or escalate difficulty (during Boredom) until the user generates effective voluntary forces.

To translate discrete firing strengths $w_k(t)$ into deterministic control policies, the policy vector $\mathbf{c}(t) = [c^*_{\text{assist}}(t), c^*_{\text{challenge}}(t)]^T$ is aggregated using a normalized weighted average defuzzification:
\begin{equation}
c^*(s^*_{\text{me}}, s^*_{\text{pe}}) = \frac{\sum_{k=1}^{N} w_k(t) \cdot O_k}{\sum_{k=1}^{N} w_k(t)}
\label{eq:defuzz}
\end{equation}
where $O_k \in [0, 1]$ represents the crisp singleton output for the $k$-th rule. By coupling Gaussian functions with Eq. (\ref{eq:defuzz}), the architecture establishes a continuously differentiable ($C^1$) control manifold. Unlike discrete Boolean triggers that cause mechanical jerk, this topology bounds the rate of change of robotic impedance, ensuring smooth force transitions despite volatile physiological inputs.

The generated control surfaces (Fig. \ref{fig:fuzzy_surface_combined}) validate this design. The assistance demand ($c^*_{\text{assist}}$, Fig. \ref{fig:fuzzy_surface_combined}(a)) peaks within the Anxiety quadrant. The steep roll-off reflects the 0.3 physiological clamp: the system provides assistance only during motor output attenuation and tapers it to zero once voluntary effort crosses the effective threshold. Conversely, the challenge level ($c^*_{\text{challenge}}$, Fig. \ref{fig:fuzzy_surface_combined}(b)) plateaus in the Flow region and extends a moderately high ridge into the Boredom quadrant. Governed by the 0.4 mental inflection point, this contour applies a progressive haptic penalty (e.g., increased stiffness) during mental distraction to prompt re-engagement.

\textit{2) Dual-Stage Decision Stabilizer:} Because physiological signals are non-stationary, evaluating the FIS outputs instantaneously can induce rapid, erratic mode switching in robotic behavior. To address this, a Dual-Stage Decision Stabilizer is introduced. First, the continuous input space is discretized into a macroscopic pending state $z_{\text{pend}}(t) \in \{\text{Flow, Anxiety, Apathy, Boredom}\}$. Rather than updating the control parameters instantaneously, a temporal dwell-time mechanism is implemented. The system updates the confirmed quadrant state $Z_k(t)$, which dictates the downstream discrete modulation factors, only if the newly decoded pending state $z_{\text{pend}}(t)$ persists continuously for a predefined threshold duration $T_{\text{dwell}}$. Any state fluctuation before exceeding $T_{\text{dwell}}$ immediately resets the internal timer. This hard reset directly rejects high-frequency physiological artifacts.

Second, a first-order low-pass filter is applied to the continuous FIS outputs ($c_{\text{assist}}^*, c_{\text{challenge}}^*$). This dual-stage stabilization ensures that macroscopic strategic mode switches ($Z_k$) remain deliberate, while microscopic parameter adjustments stay continuous and safe. Together, these stabilized outputs form the complete high-level control commands sent to the physical execution layer.

\section{Adaptive Impedance Control}
The low-level controller (Fig. \ref{fig:quadrant_model}(b)) translates the strategic intent derived from the high-level engine into physical robotic interaction via a Variable Impedance Control scheme. This architecture enables sensorless compliance modulation, ensuring intrinsic safety while executing the prescribed therapeutic strategies.

\subsection{Adaptive Parameter Synthesis}
This module synthesizes the continuous policy variables ($c_{\text{assist}}^*, c_{\text{challenge}}^*$) and the macroscopic state ($Z_k$) into actionable physical parameters: stiffness $K(t)$, damping $D(t)$, and assistance force $F_{\text{assist}}(t)$. 

To govern the user's control authority, two state-dependent modulation factors are introduced, specifically the Strength Reduction Factor ($\gamma_{\text{str}}$) and the Co-activation Penalty Factor ($\delta_{\text{co}}$):
\begin{equation}
\gamma_{\text{str}}(Z_k) = \delta_{\text{co}}(Z_k) =
\begin{cases}
1.0, & \text{if } Z_k = Z_{\text{Flow}} \\
0.5, & \text{if } Z_k = Z_{\text{Boredom}} \\
0.0, & \text{if } Z_k \in \{Z_{\text{Anxiety}}, Z_{\text{Apathy}}\}
\end{cases}
\end{equation}
This discrete allocation ensures safety. Factors of 1.0 in the Flow state maximize voluntary authority, while factors of 0.0 during Anxiety or Apathy isolate the control loop from erratic signals, transferring full authority to the robot for stable guidance.

Based on these factors, the assistive force $F_{\text{assist}}(t)$ is formulated as:
\begin{equation}
F_{\text{assist}}(t) = \text{sgn}(d(t)) \cdot \left(F_{\text{min}} + c^*_{\text{assist}}(t)\Delta F\right) \cdot \left(1 - \gamma_{\text{str}}(Z_k)s^*_{\text{pe}}\right)
\end{equation}
where $\Delta F = F_{\text{max}} - F_{\text{min}}$ and $d(t)$ is the desired direction. This multiplicative fusion ensures zero-latency assistance withdrawal. Even under high commanded assistance, any spontaneous valid effort ($s^*_{\text{pe}}$) immediately attenuates the robotic power output, achieving Assistance-As-Needed (AAN) behavior.

The impedance parameters ($K$ and $D$) are directly mapped from the challenge policy:
\begin{align}
K(t) &= K_{\text{min}} + c^*_{\text{challenge}}(t) \cdot \Delta K \\
D(t) &= (D_{\text{min}} + c^*_{\text{challenge}}(t) \cdot \Delta D) \cdot \left(1 + \delta_{\text{co}}(Z_k)\beta_{\text{co}}s_{\text{co}}(t)\right)
\end{align}
where $\Delta K$ and $\Delta D$ represent the respective parameter ranges, and $\beta_{\text{co}}$ is a scaling coefficient. $K(t)$ determines the strictness of the spatial constraint, while $D(t)$ amplifies baseline damping proportionally to the co-activation level $s_{\text{co}}(t)$. This haptic penalty makes excessive co-contraction feel artificially ``heavy,'' driving the user to adopt relaxed motor synergies.

Although the modulation factors ($\gamma_{\text{str}}$ and $\delta_{\text{co}}$) undergo discrete step changes, they do not compromise continuous physical interaction. This smoothness is guaranteed through two mechanisms. First, the Dual-Stage Stabilizer (Section III-C.2) strictly bounds the transition frequency. Second, a low-level Exponential Moving Average (EMA) filter is applied to bridge these targets with mechanical execution. Let $\Phi \in \{K, D, F_{\text{assist}}\}$ denote the synthesized target parameters. The actual commanded parameter $\Phi_{\text{cmd}}(t)$ is updated via:
\begin{equation}
\Phi_{\text{cmd}}(t) = \alpha \cdot \Phi_{\text{cmd}}(t-\Delta t) + (1 - \alpha) \cdot \Phi(t)
\end{equation}
where $\alpha \in [0, 1)$ is the smoothing factor ($\alpha=0.8$ at 100 Hz). Combined with the continuously differentiable cognitive policies, this kinetic filtering translates macroscopic mode switches into dynamically smooth haptic interventions without mechanical jerk.

\subsection{Variable Impedance Control}
To ensure intrinsic safety and robustness against sensor noise in dynamic settings, a position-based Variable Impedance Control (VIC) scheme is employed. This approach modulates interaction compliance directly, without relying on potentially unstable force-feedback loops.

The actuator command torque $\tau_{\text{cmd}}$ is formulated as a superposition of feedback impedance, feedforward assistance, and static friction compensation \cite{moca}:
\begin{equation}
\tau_{\text{cmd}} = K(t)(q_d - q) + D(t)(\dot{q}_d - \dot{q}) + \tau_{\text{assist}}(t) + \tau_{\text{fric}}
\end{equation}
where $q$ and $q_d$ denote the actual and desired joint positions, respectively. By structurally decoupling the time-varying feedback impedance ($K(t)$ and $D(t)$) from the feedforward assistance ($\tau_{\text{assist}}$, derived from $F_{\text{assist}}(t)$ via the Jacobian transpose), the low-level controller independently executes the multidimensional AAN interventions directed by the high-level cognitive strategies.

\section{Experiment and Characteristic Evaluation}
Based on the proposed framework, experimental validations were conducted to assess its efficacy across diverse mental and physical states. As illustrated in Fig. \ref{fig:protocol}, the overall experimental procedure comprised an individualized calibration phase followed by two primary validation studies.

\begin{figure}[htbp]
    \centering
    \captionsetup[subfloat]{captionskip=2pt} 
    
    \subfloat[\label{fig:surf_assist}]{
        \includegraphics[width=0.27\linewidth]{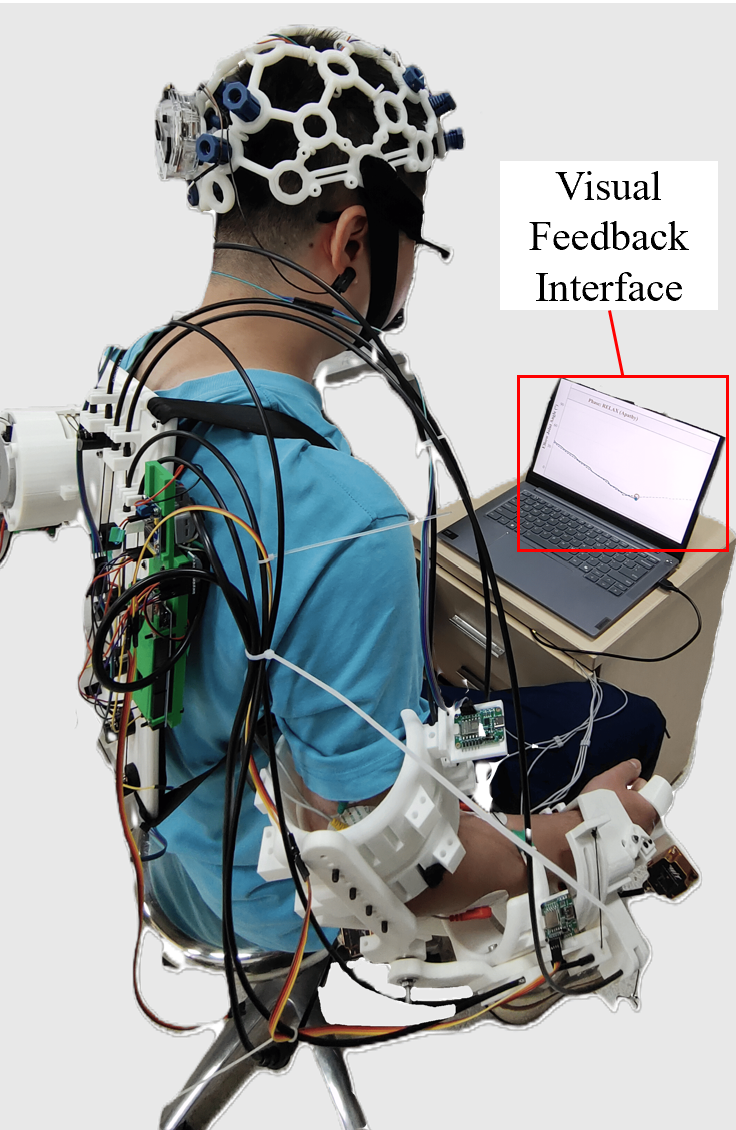}
    }
    \hfill 
    \subfloat[\label{fig:surf_challenge}]{
        \includegraphics[width=0.67\linewidth]{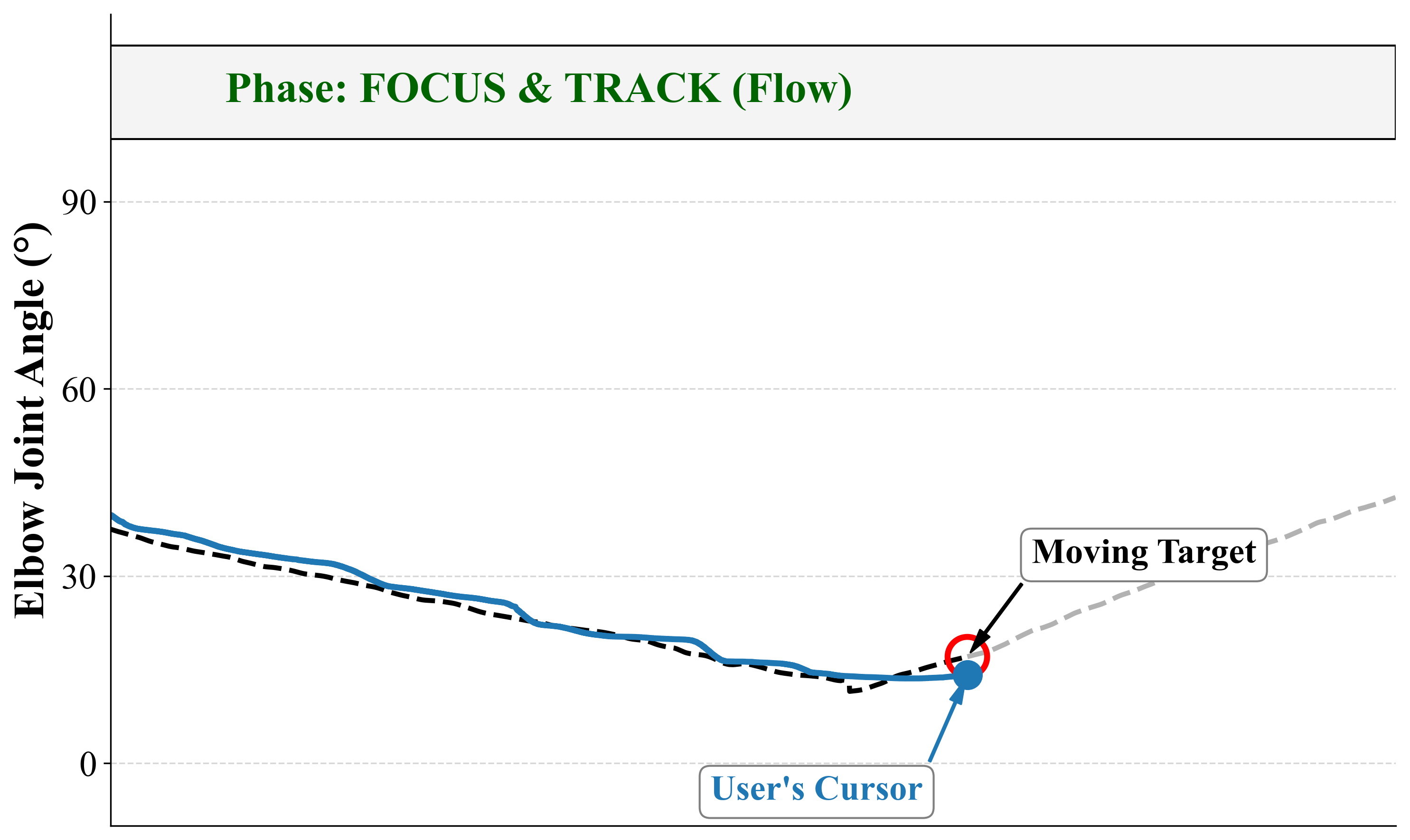}
    }
    
    \caption{Experimental setup and the visual feedback interface. (a) Overview of the experimental scenario during the active tracking task. (b) A representative screenshot of the custom graphical user interface (GUI), illustrating the target trajectory, the user's real-time cursor, and the synchronized phase prompts.}
    \label{fig:experiment_protocol}
\end{figure}

\subsection{Experimental Participants and Setup}
Ten healthy, right-handed volunteers (age $23.3 \pm 1.8$ years; 8 males, 2 females) participated in the study under institutional ethics approval (no. 20250066). The CURE upper-limb exoskeleton was utilized for 1-DoF elbow flexion/extension. As illustrated in Fig. \ref{fig:experiment_protocol}(a), seated participants interacted with a custom visual feedback interface on a monitor. Through this interface (Fig. \ref{fig:experiment_protocol}(b)), participants tracked a continuous sinusoidal target (0.2 Hz, range $0^\circ$--$90^\circ$) by controlling a real-time cursor that reflected their actual elbow joint angle. The interface also delivered synchronized on-screen text and audio cues to guide subjects through specific behavioral phases. To emulate the elevated physical demand and intent-execution mismatch often encountered in robot-assisted training, a 1.2 kg weight was affixed to the user's wrist. This added inertial load ensures sustained muscular effort and natural fatigue during prolonged tracking tasks.

As depicted in Fig. \ref{fig:protocol}(a), an individualized calibration session was conducted prior to the main trials. The normalization bounds ($[s_{\text{min}}, s_{\text{max}}]$) for the EEG decoder were established through a 60-second resting baseline, a passive following task (eyes closed, mental relaxation), and an active tracking task (maximal visual focus). Concurrently, physical engagement for the sEMG decoder was normalized against peak values obtained from a 5-second Maximum Voluntary Contraction (MVC) test.

\subsection{Experiment Protocol}

\begin{figure}[t]
    \centering
    \subfloat[]{
        \includegraphics[width=0.95\linewidth]{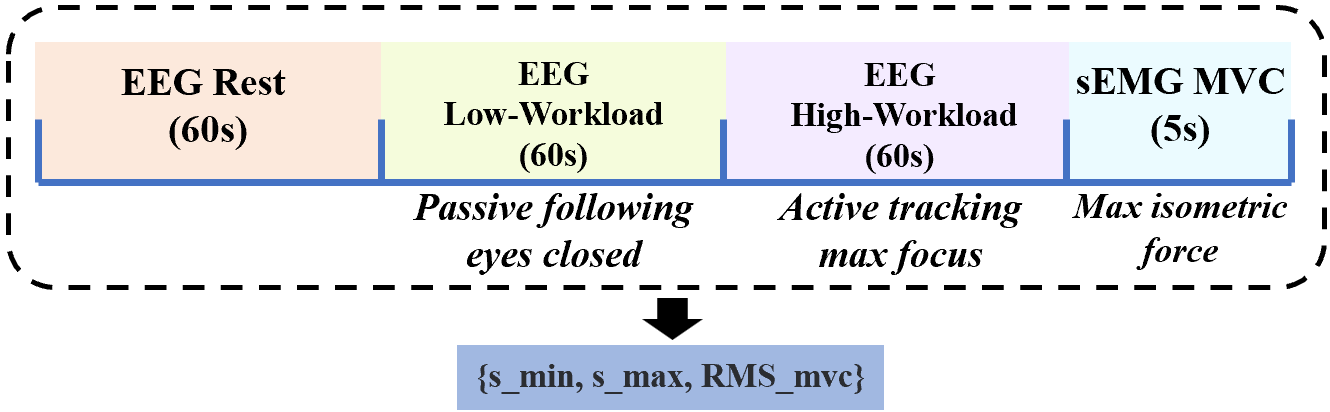}
        \label{fig:sub1}
    } \\ 
    \vspace{-0.2cm} 
    \subfloat[]{
        \includegraphics[width=0.95\linewidth]{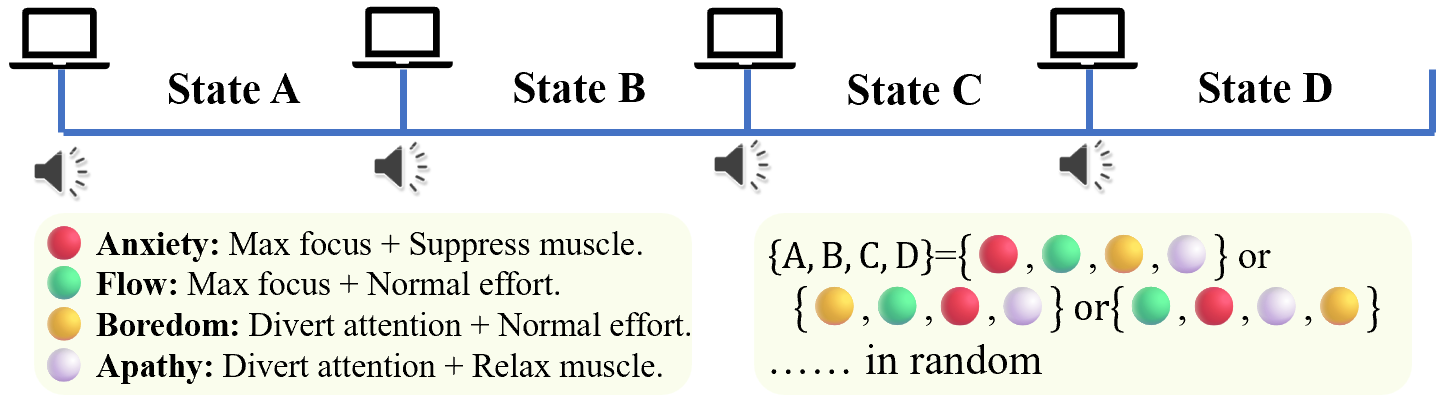}
        \label{fig:sub2}
    } \\ 
    \vspace{-0.2cm}
    \subfloat[]{
        \includegraphics[width=0.95\linewidth]{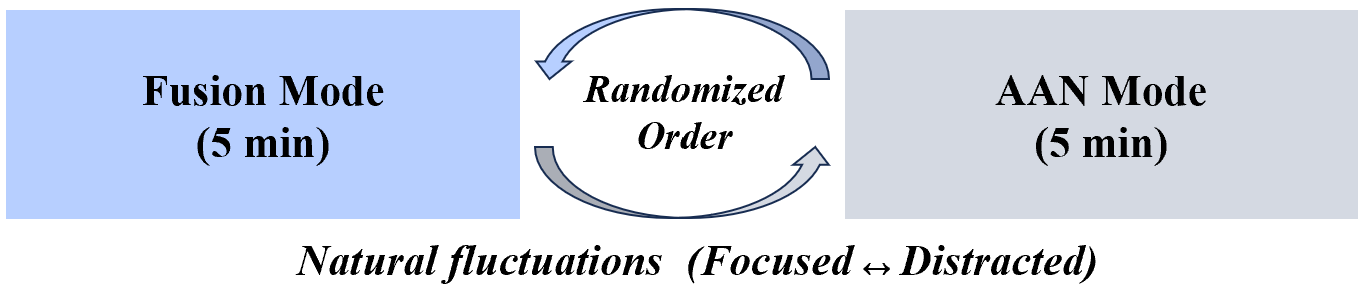}
        \label{fig:sub3}
    } \\ 
    \vspace{-0.2cm}
   \caption{Experimental procedure and protocol design. (a) Individualized calibration phase. (b) Study I: Continuous adaptive response test (240-s). (c) Study II: Comparative efficacy analysis against a baseline AAN mode (5-min).}
    \label{fig:protocol}
\end{figure}

As illustrated in Fig.~\ref{fig:protocol}, the experimental validation was structured into three sequential phases to establish physiological baselines, characterize the system's dynamic stability, and evaluate its functional efficacy. 

\textit{1) Preliminary Parameter Tuning (Fig.~\ref{fig:protocol}(a)):} To empirically establish the Gaussian membership inflection points ($c=0.4$ and $0.3$) for the decision engine, participants completed 1-minute isolated trials. Maximal and minimal mental engagements were induced via active visual tracking versus distracted tracking. Correspondingly, physical engagement boundaries were established using a ``Zero-Assist'' mode (requiring maximal voluntary effort) versus a ``Passive Following'' mode (complete arm relaxation). 

\textit{2) Continuous Adaptive Response Test (Fig.~\ref{fig:protocol}(b)):} To evaluate real-time functional responses to specific anomalies (e.g., intent-execution mismatch), participants completed a continuous 240-s session. Guided by audio-visual cues, they traversed the four Flow quadrants at 60-s intervals. Specific behaviors were safely induced: intense focus with active tracking (Flow), visual distraction with physical tracking (Boredom), intense focus with intentionally suppressed muscular output (Anxiety), and total relaxation (Apathy). System stability during these pseudo-randomized state transitions was quantified using Transition Latency and Maximum Force Derivative ($|dF_{\text{assist}}/dt|$) to ensure safe, chatter-free interventions.

\textit{3) Comparative Efficacy Analysis (Fig.~\ref{fig:protocol}(c):} To quantitatively assess the multimodal architecture, participants performed three 5-minute sessions under the Fusion Mode (FM) and a baseline EMG-driven AAN mode. This baseline represents current slacking-prevention protocols \cite{emg_aan1, emg_aan4}, which decay feedforward assistance proportionally to physical engagement ($F_{\text{assist}} \propto 1 - s^*_{\text{pe}}$) while maintaining constant guiding stiffness. Performance was assessed via kinematic tracking error (RMSE) and the raw robotic assistive force ($|F_{\text{assist}}|$). To prevent circular reasoning when evaluating physiological exertion, high-fidelity offline metrics (iEMG and EEG PSD) were extracted using zero-phase Butterworth filtering. This non-causal processing eliminates phase distortion, providing an independent physiological ground truth to quantify the Flow State Ratio.

\subsection{Experimental Results}

\textit{1) Results of Preliminary Parameter Tuning:}
To empirically justify the fuzzy membership parameterization (Section III-C), the steady-state physiological responses from the 1-minute isolated tuning blocks were analyzed. As summarized in Table \ref{tab:pilot_tuning}, the specific task conditions (e.g., ``Zero-Assist'' vs. ``Passive Following'', and active focus vs. distraction) elicited distinct and statistically separable physiological boundaries across all 10 participants.

\begin{table}[htbp]
    \centering
    \caption{Physiological Responses During the Isolated 1-Minute Induction Blocks (Mean $\pm$ SD)}
    \label{tab:pilot_tuning}
    \renewcommand{\arraystretch}{1.3}
    \resizebox{\columnwidth}{!}{%
    \begin{tabular}{lccc}
    \toprule
    \textbf{Physiological Metric} & \textbf{Induced ``Low'' State} & \textbf{Induced ``High'' State} & \textbf{Selected Inflection ($c$)} \\
    \midrule
    Mental Engagement ($s^*_{\text{me}}$)  & $0.22 \pm 0.08$ & $0.65 \pm 0.12$ & \textbf{0.40} \\
    Physical Engagement ($s^*_{\text{pe}}$) & $0.15 \pm 0.06$ & $0.58 \pm 0.10$ & \textbf{0.30} \\
    \bottomrule
    \end{tabular}%
    }
\end{table}

These empirical distributions reveal that a mathematically symmetric $0.5$ threshold would significantly overlap with the lower confidence intervals of the ``High'' states, leading to a punitive control scheme. By anchoring the Gaussian inflection points at $0.3$ for $s^*_{\text{pe}}$ and $0.4$ for $s^*_{\text{me}}$ (Table~\ref{tab:pilot_tuning}), the 95\% confidence intervals of both distributions are separated. This data-driven thresholding allows the fuzzy engine to delineate active voluntary effort from passive slacking or mental disengagement.

\begin{figure}[htbp]
    \centering
    \captionsetup[subfloat]{captionskip=0pt} 
    
    \subfloat[]{\includegraphics[width=\linewidth]{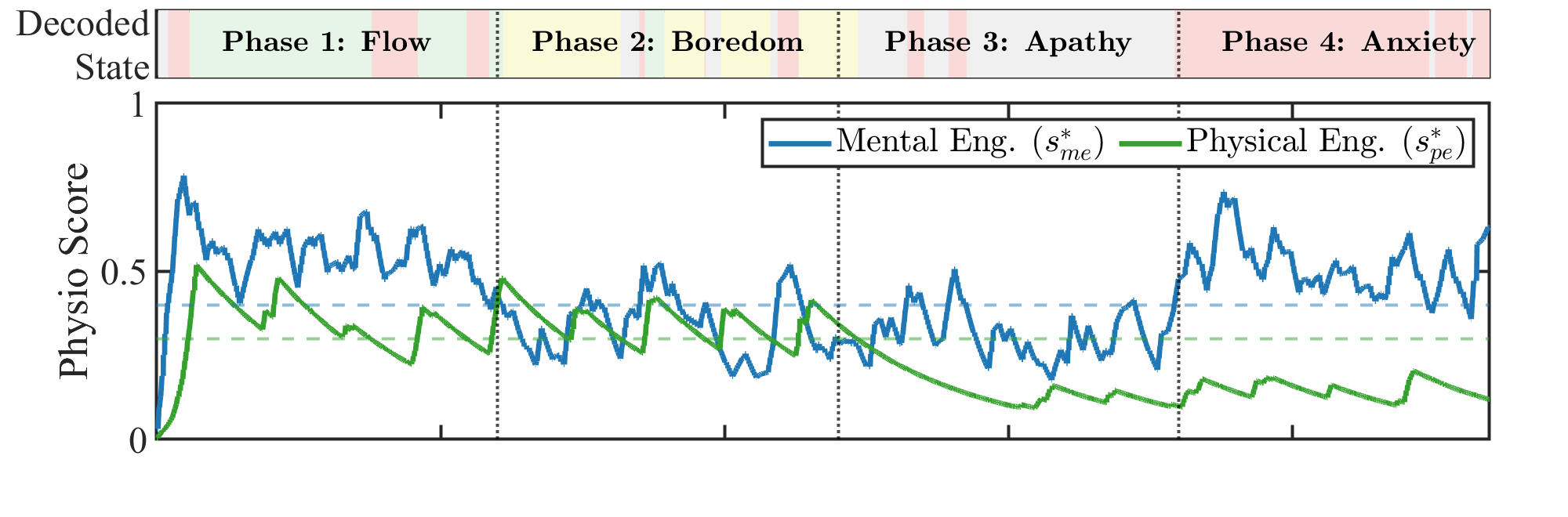}\label{fig:sub1}} \\ 
    \vspace{-0.2cm} 
    
    \subfloat[]{\includegraphics[width=\linewidth]{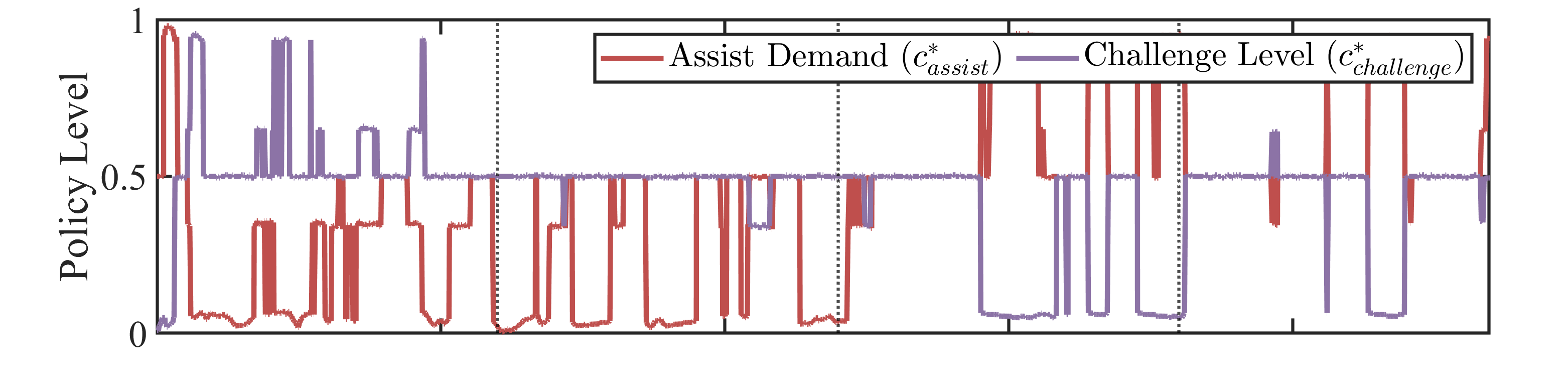}\label{fig:sub2}} \\ 
    \vspace{-0.2cm}
    
    \subfloat[]{\includegraphics[width=\linewidth]{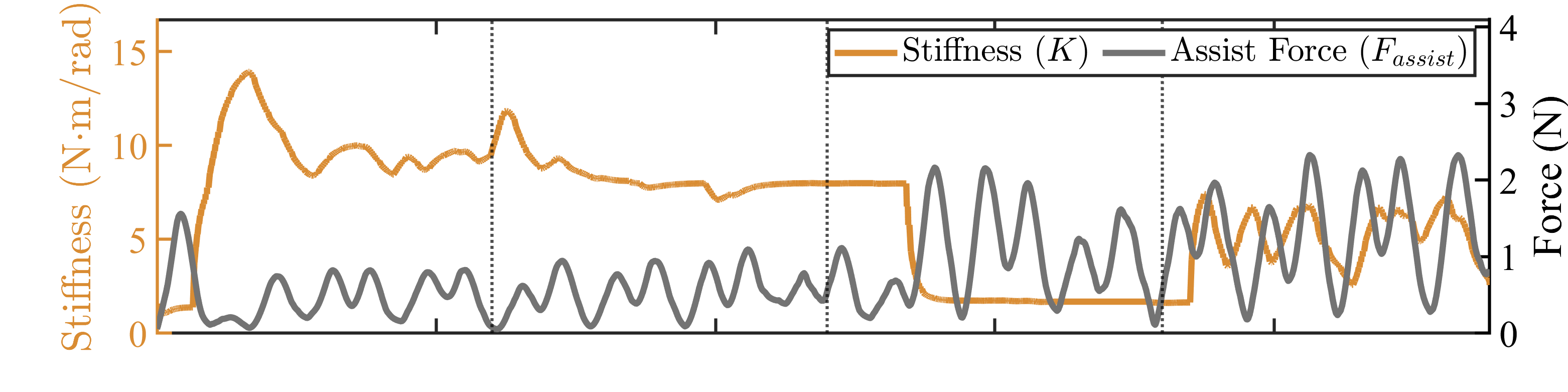}\label{fig:sub3}} \\ 
    \vspace{-0.2cm}
    
    \subfloat[]{
        \captionsetup[subfloat]{captionskip=7pt}
        \includegraphics[width=\linewidth]{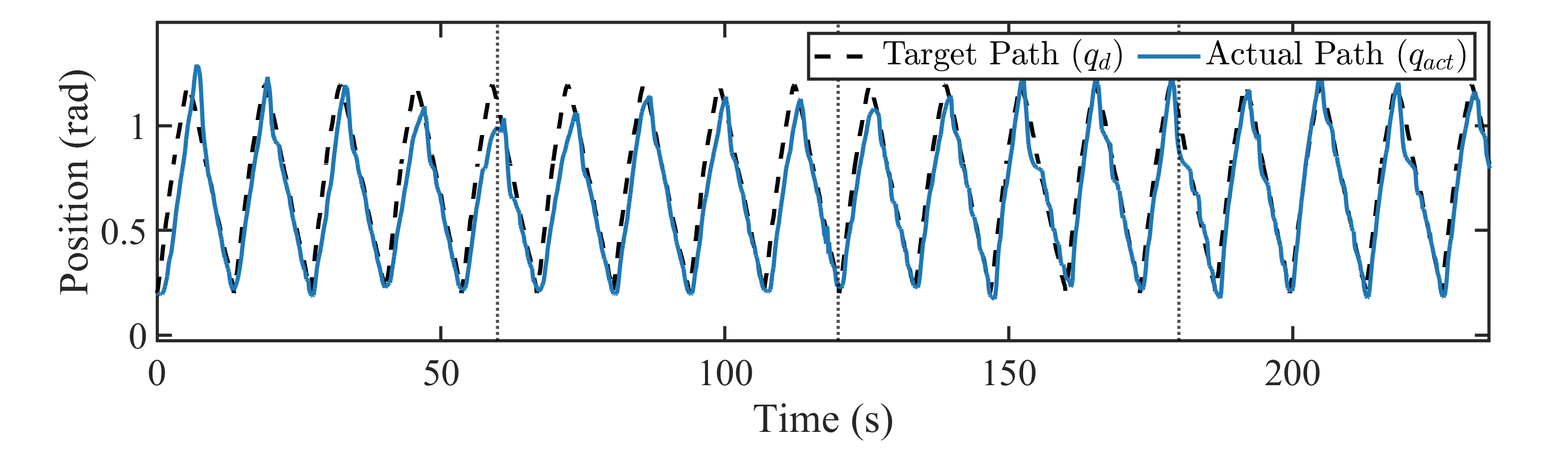}\label{fig:sub4}
    }\\
    \vspace{-0.2cm} 
    
    \caption{System dynamic response across the four quadrants during a continuous 240-s training session. The top ribbon indicates the real-time decoded macroscopic states ($Z_k$). (a) Smoothed physiological inputs: mental engagement ($s^*_{\text{me}}$) and physical engagement ($s^*_{\text{pe}}$). (b) High-level continuous policy variables: assistance demand ($c^*_{\text{assist}}$) and challenge level ($c^*_{\text{challenge}}$). (c) Low-level modulated impedance parameters: virtual stiffness ($K$) and feedforward assistance force ($F_{\text{assist}}$). (d) Kinematic tracking performance.}
    \label{fig:four_vertical_figures}
\end{figure}

\textit{2) Results of Continuous Adaptive Response Test:}
Fig.~\ref{fig:four_vertical_figures} illustrates a representative 240-s session from Subject 6, with the top ribbon indicating the real-time decoded macroscopic states. Table~\ref{tab:phase_response} summarizes the specific robotic interventions extracted from the actual trajectories across the four induced phases.

\begin{table}[htbp]
    \centering
    \caption{Summary of System Responses Across the Four Induced States}
    \label{tab:phase_response}
    \renewcommand{\arraystretch}{1.2}
    \resizebox{\columnwidth}{!}{%
    \begin{tabular}{lcccc}
    \toprule
    \textbf{Phase (State)} & $\mathbf{s^*_{\text{me}}}$ & $\mathbf{s^*_{\text{pe}}}$ & \textbf{Stiffness $K$ (N$\cdot$m/rad)} & \textbf{Assist. Force $F_{\text{assist}}$ (N)} \\
    \midrule
    1 (Flow) & High & High & High ($\sim 10$) & Minimal ($<1.0$) \\
    2 (Boredom) & Low & High & Firm ($\sim 8$) & Withheld ($<1.0$) \\
    3 (Apathy) & Low & Low & Low ($\sim 2$) & Moderate Peaks ($\sim 2.0$) \\
    4 (Anxiety) & High & Low & Fluctuating ($\sim 5$) & Strong Peaks ($\sim 2.3$) \\
    \bottomrule
    \end{tabular}%
    }
\end{table}

As detailed in Table~\ref{tab:phase_response}, the system tailors its interventions across all four psychological states. In the Flow phase, the robot minimizes assistance to maximize voluntary effort while maintaining a firm guiding stiffness. During Boredom, it continues to withhold power and enforces a haptic constraint to demand mental re-engagement. When the user disengages (Apathy), the controller prioritizes safety by reducing the stiffness ($K \sim 2\,\text{N}\cdot\text{m/rad}$) to prevent rigid dragging. Conversely, during an intent-execution mismatch (Anxiety), the system injects feedforward assistance to ensure trajectory completion (Fig.~\ref{fig:four_vertical_figures}(c), (d)).

\begin{figure}[htbp]
    \centering
    \subfloat[]{
        \includegraphics[width=0.46\linewidth]{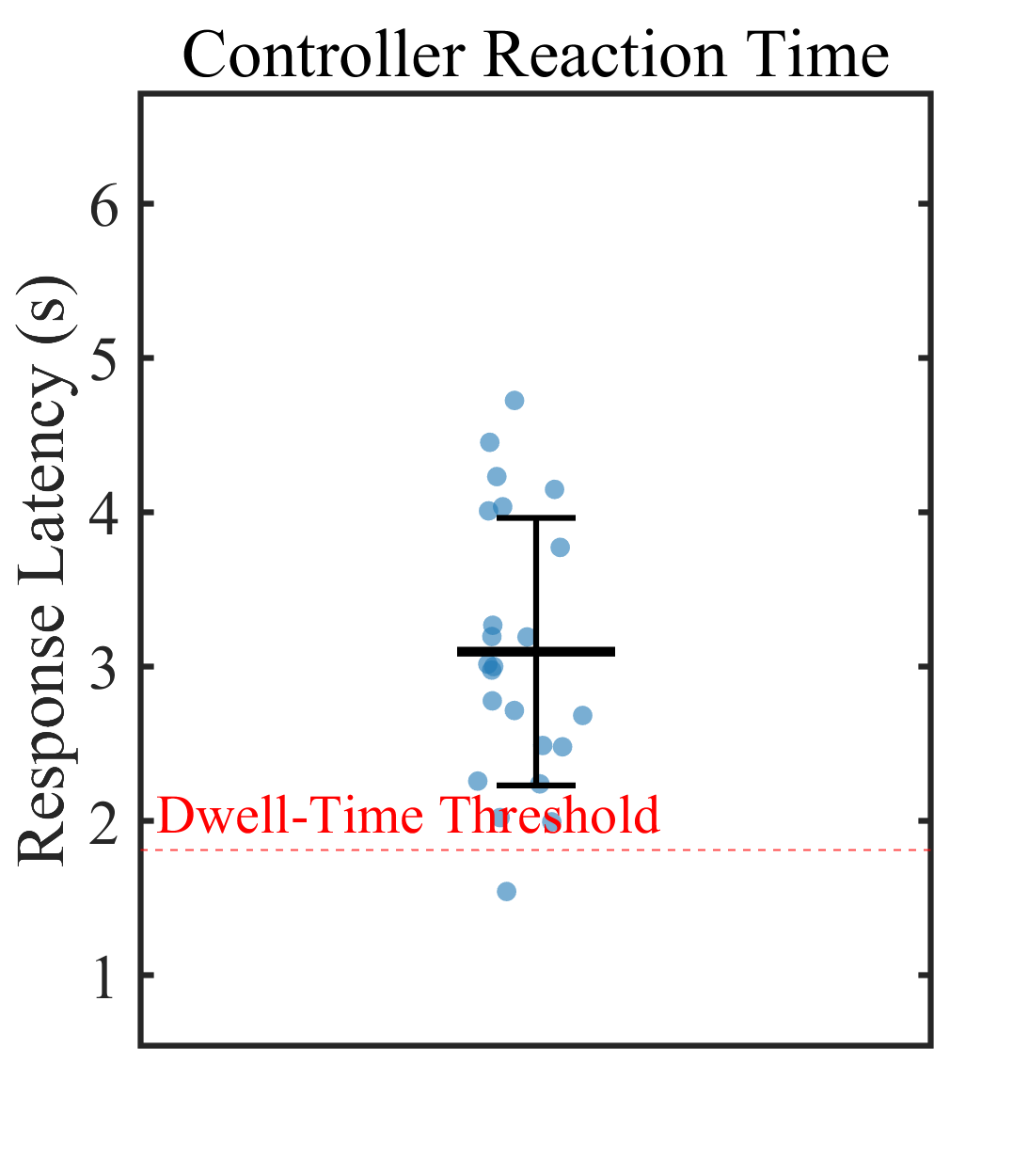} 
        \label{fig:stats_latency}
    }
    \hfill
    \subfloat[]{
        \includegraphics[width=0.46\linewidth]{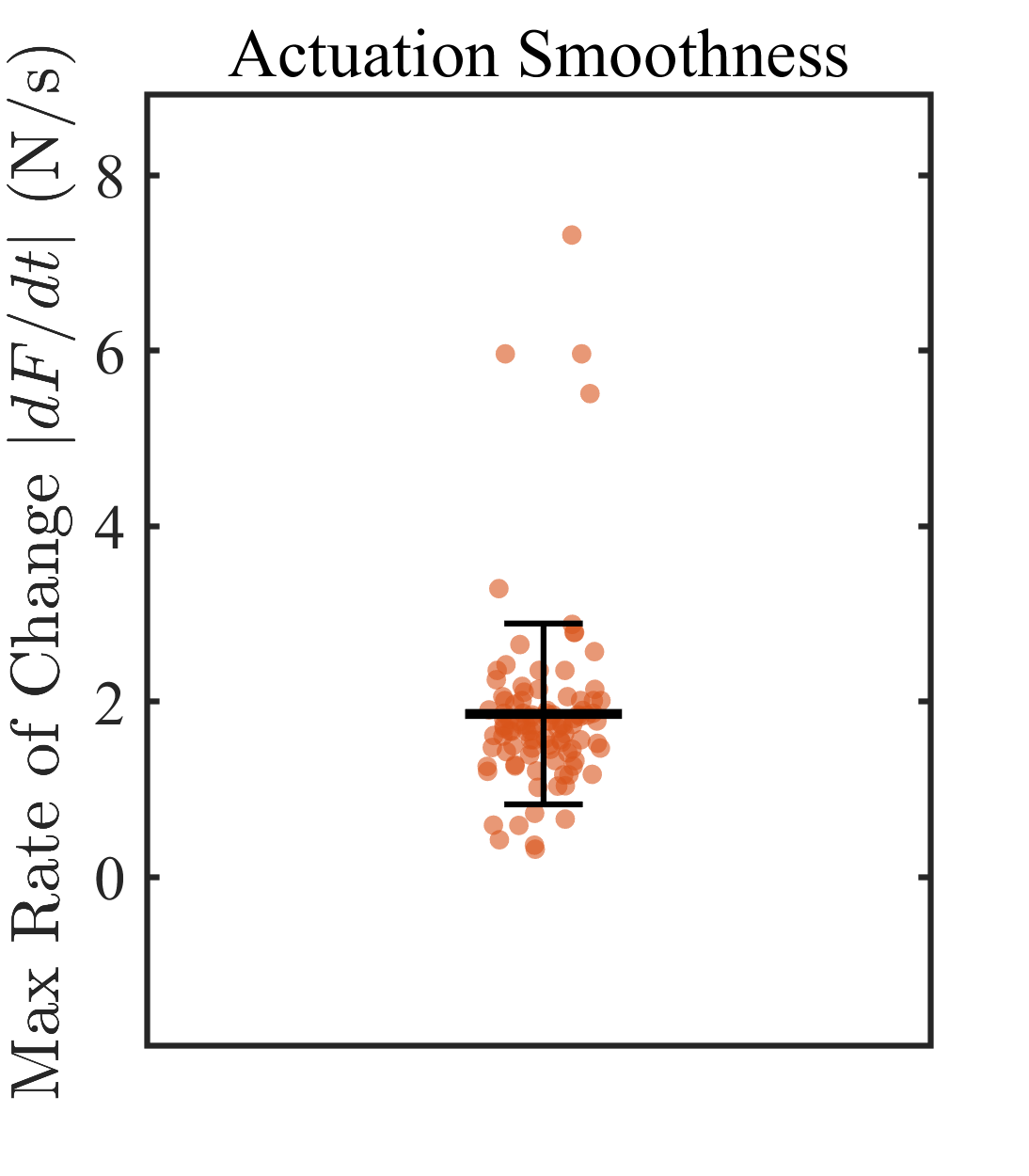} 
        \label{fig:stats_jerk}
    }
    \caption{Quantitative analysis of transition smoothness across 23 major inter-quadrant shifts. (a) Boxplot of the controller response latency. (b) Boxplot of the maximum rate of change of the assistance force ($|dF_{\text{assist}}/dt|$).}
    \label{fig:transition_stats}
\end{figure}

To quantify the transition dynamics, 23 major inter-quadrant shifts were evaluated (Fig.~\ref{fig:transition_stats}). The average transition latency ($3.10 \pm 0.87\,\text{s}$) matches the combined Dwell-Time and low-pass filter delay. Furthermore, the macroscopic rate of change of the assistance force ($|dF_{\text{assist}}/dt|$) averaged $1.86 \pm 1.03\,\text{N/s}$, remaining well within comfortable bounds. These metrics confirm that the system translates macroscopic state shifts into smooth, chatter-free haptic interventions.

\textit{3) Results of Comparative Efficacy Analysis:}
The FM demonstrated significant advantages over the baseline AAN mode across all participants, as summarized in Table~\ref{tab:efficacy_comparison}.

\begin{table}[htbp]
    \centering
    \caption{Quantitative Comparison Between Baseline AAN and Proposed Fusion Mode (FM)}
    \label{tab:efficacy_comparison}
    \renewcommand{\arraystretch}{1.2}
    \resizebox{\columnwidth}{!}{%
    \begin{tabular}{lccc}
    \toprule
    \textbf{Evaluation Metric} & \textbf{Baseline AAN} & \textbf{Fusion Mode (FM)} & \textbf{\textit{p}-value} \\
    \midrule
    Tracking RMSE (rad) & $0.177$ & $\mathbf{0.174}$ & $> 0.05$ (n.s.) \\
    Mean Assistive Force (N) & $1.85$ & $\mathbf{0.51}$ & $< 0.001$ \\
    Mental Engagement ($s^*_{\text{me}}$) & $0.35$ & $\mathbf{0.45}$ & $< 0.01$ \\
    Physical Engagement ($s^*_{\text{pe}}$) & $0.21$ & $\mathbf{0.34}$ & $< 0.001$ \\
    Flow State Ratio (\%) & $4.6$ & $\mathbf{42.8}$ & $< 0.001$ \\
    \bottomrule
    \end{tabular}%
    }
\end{table}

\begin{figure}[htbp]
    \centering
    \subfloat[]{
        \includegraphics[width=\linewidth]{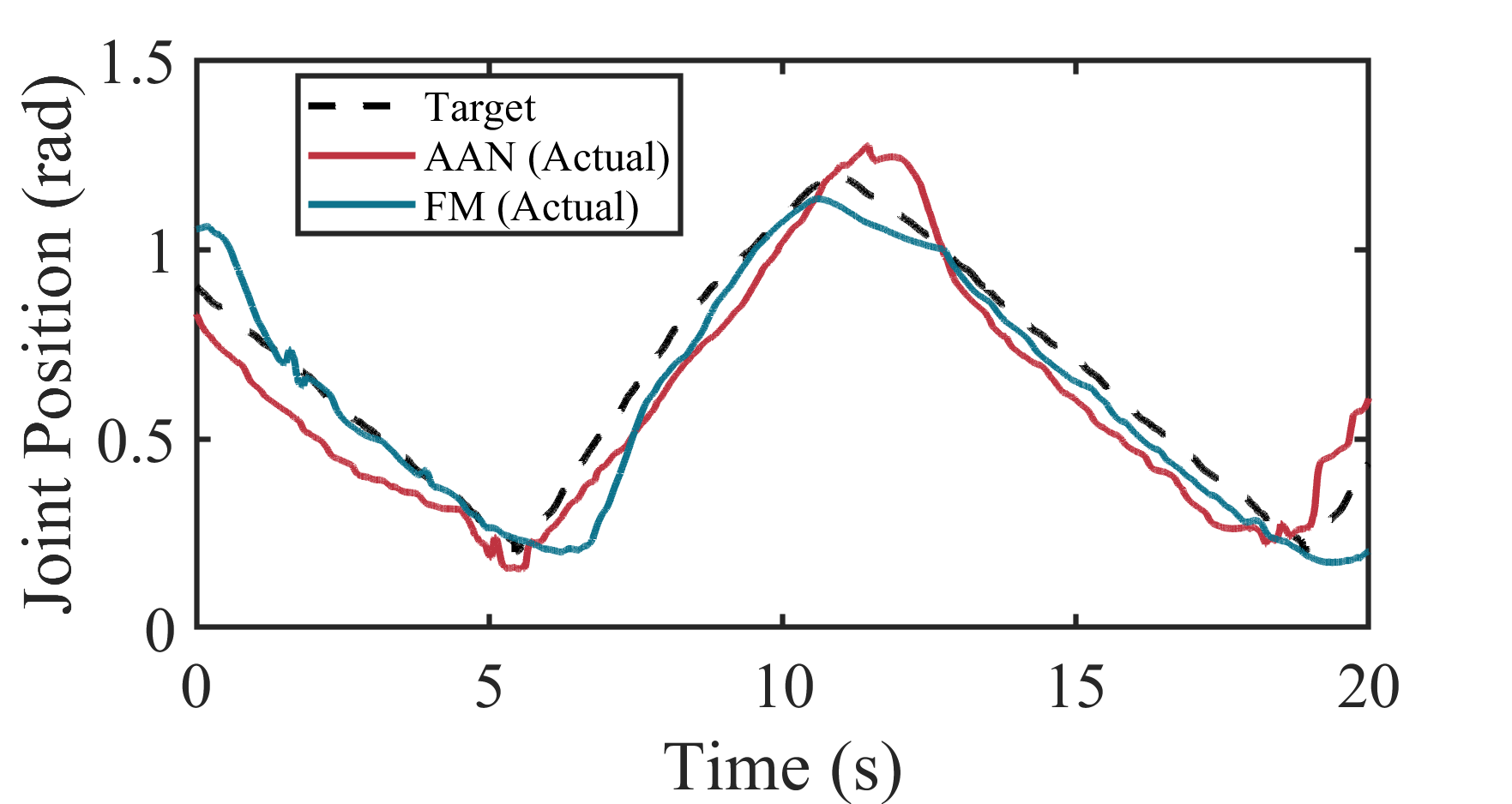}
        \label{fig:traj_sample}
    } \\ 
    \subfloat[]{
        \includegraphics[width=0.45\linewidth]{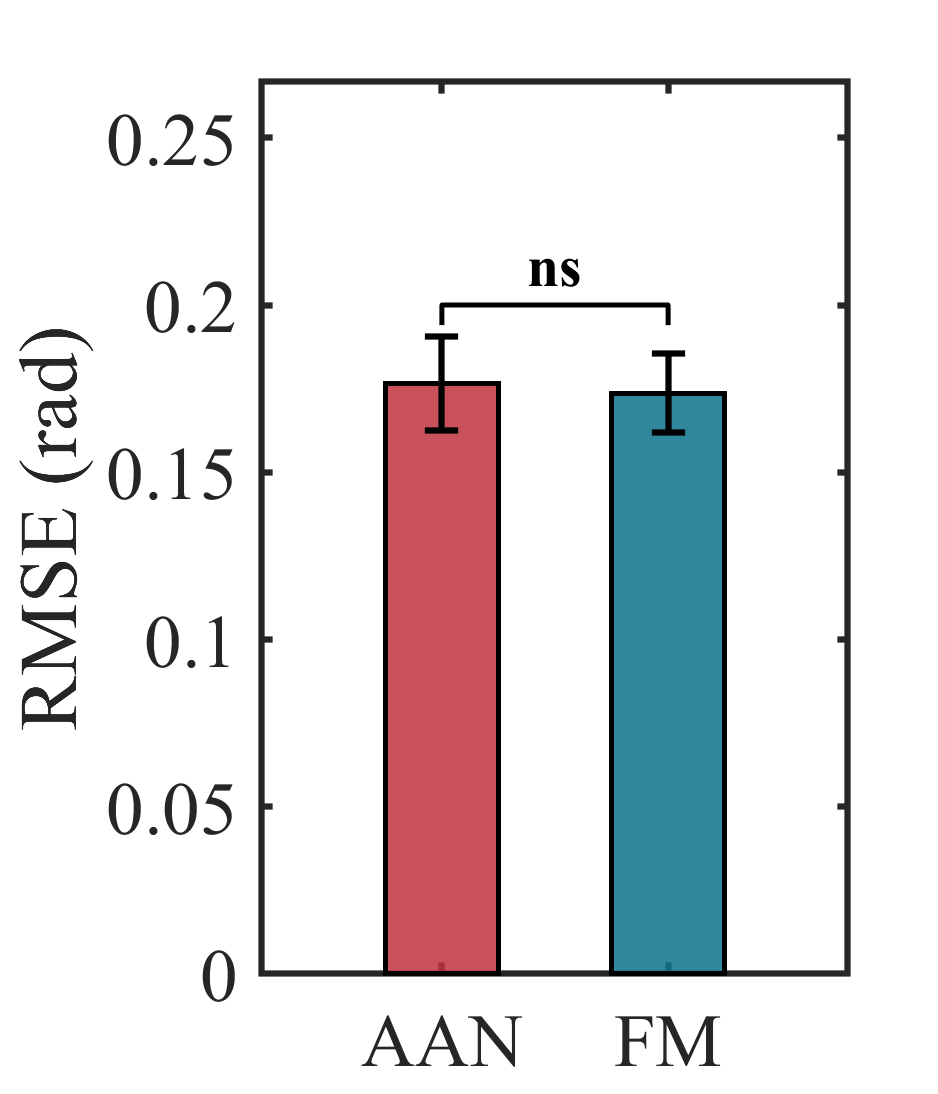}
        \label{fig:rmse_bar}
    }
    \hfill
    \subfloat[]{
        \includegraphics[width=0.45\linewidth]{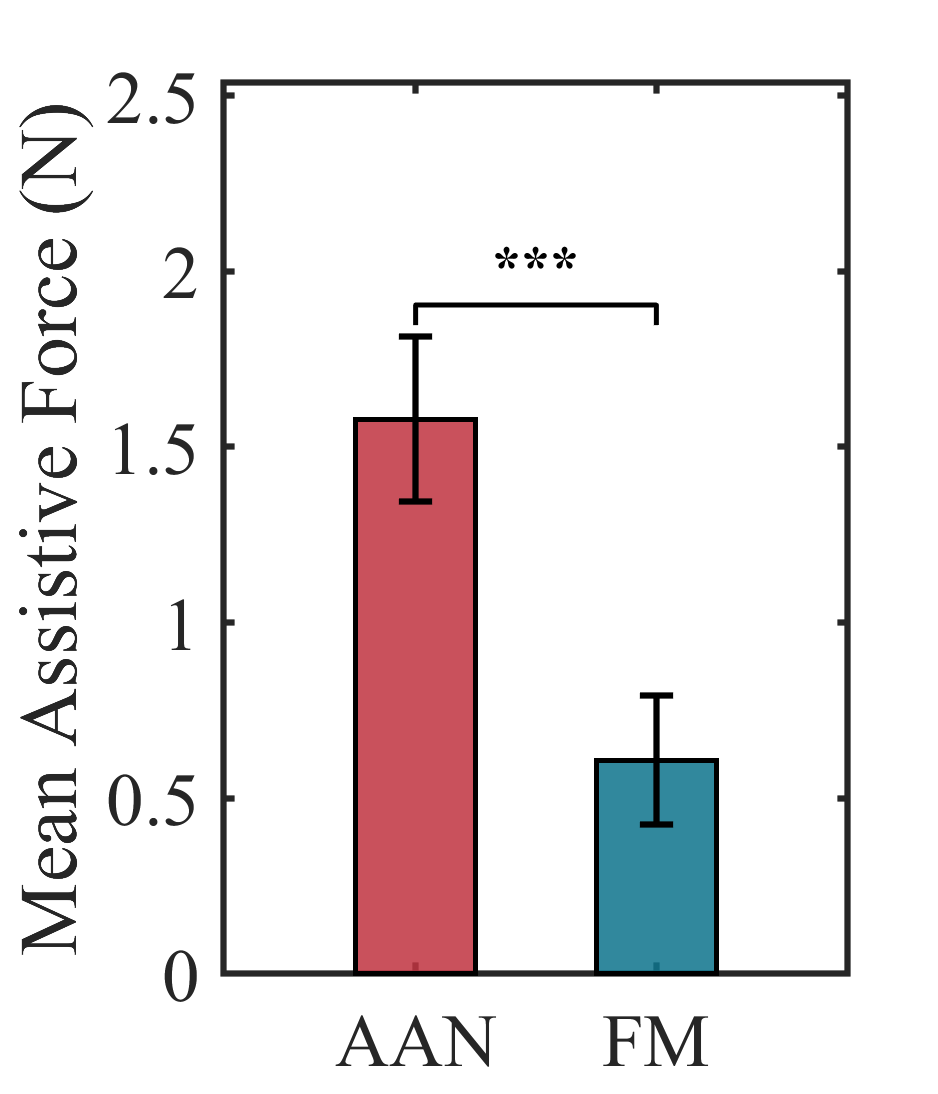}
        \label{fig:power_bar}
    }
    \caption{Kinematic tracking performance and robotic assistance evaluation. (a) A representative tracking trajectory segment under AAN and Fusion Mode (FM). (b) Grand-averaged Root Mean Square Error (RMSE) across all subjects ($p > 0.05$). (c) Grand-averaged mean feedforward assistive force showcasing a significant reduction in robotic physical intervention ($p < 0.001$).}
    \label{fig:macro_performance}
\end{figure}

\begin{figure}[htbp]
    \centering

    \subfloat[]{
        \includegraphics[width=\linewidth]{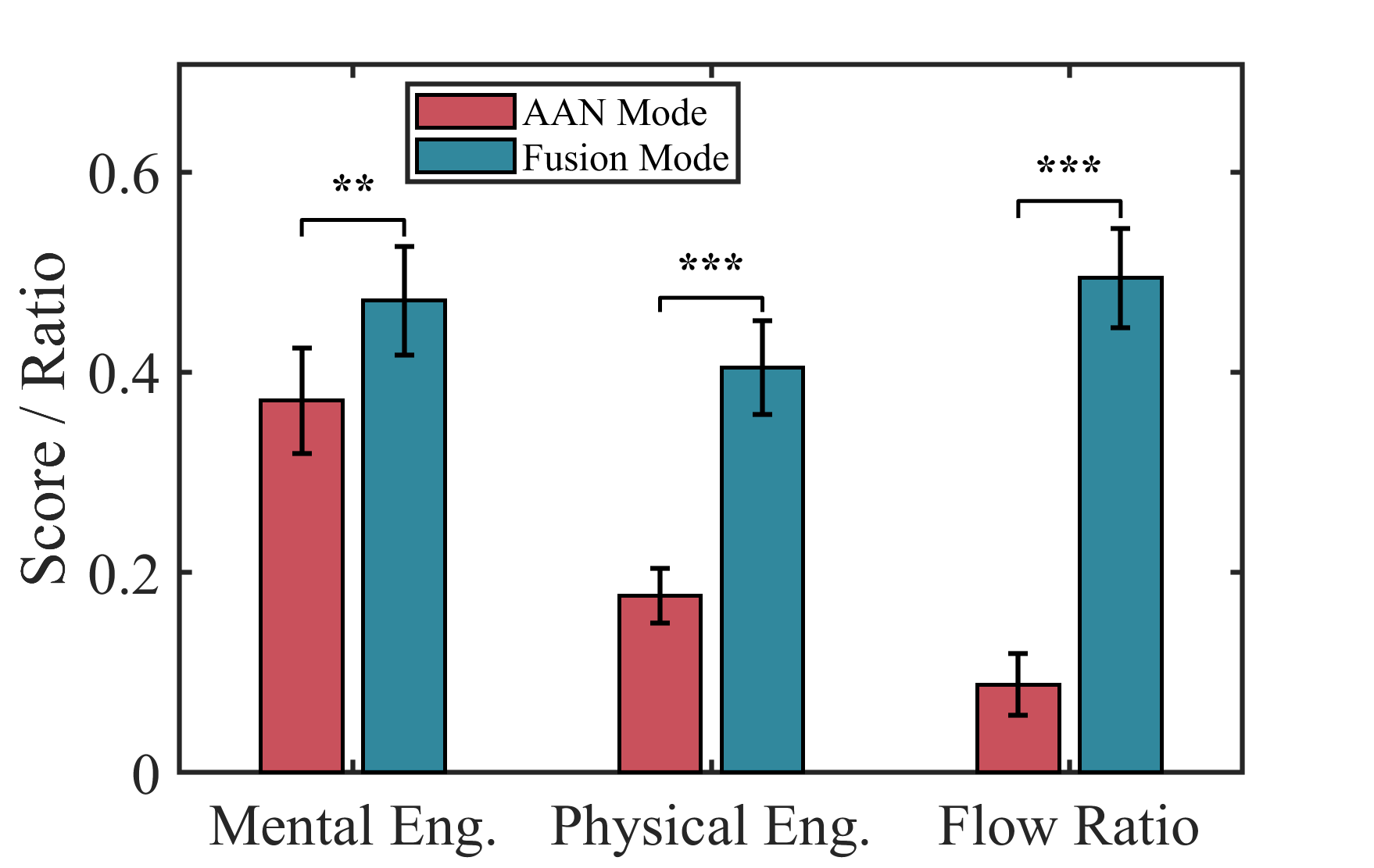}
        \label{fig:physio_scores}
    } \\ 

    \subfloat[]{
        \includegraphics[width=0.47\linewidth]{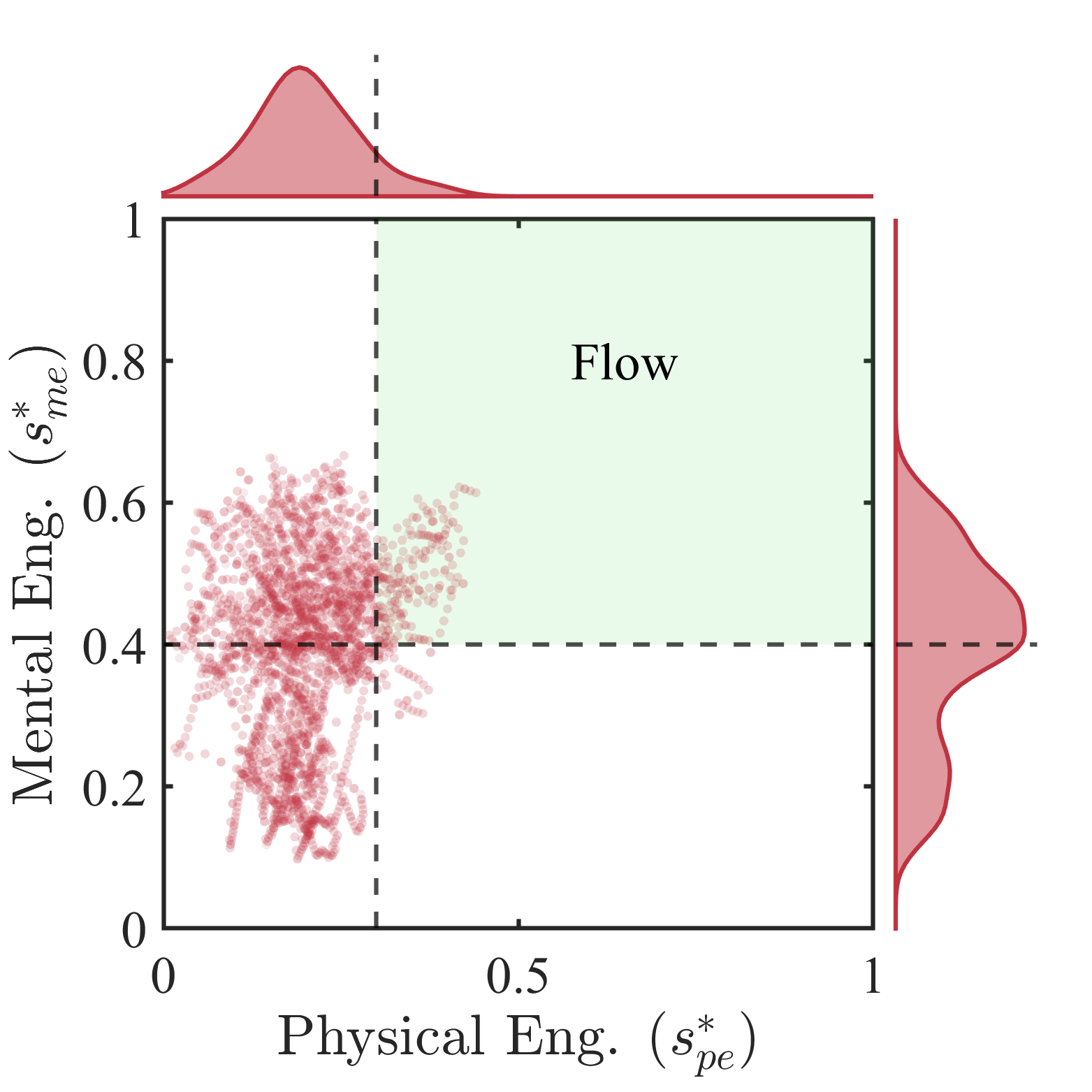}
        \label{fig:kde_aan}
    }
    \hfill
    \subfloat[]{
        \includegraphics[width=0.47\linewidth]{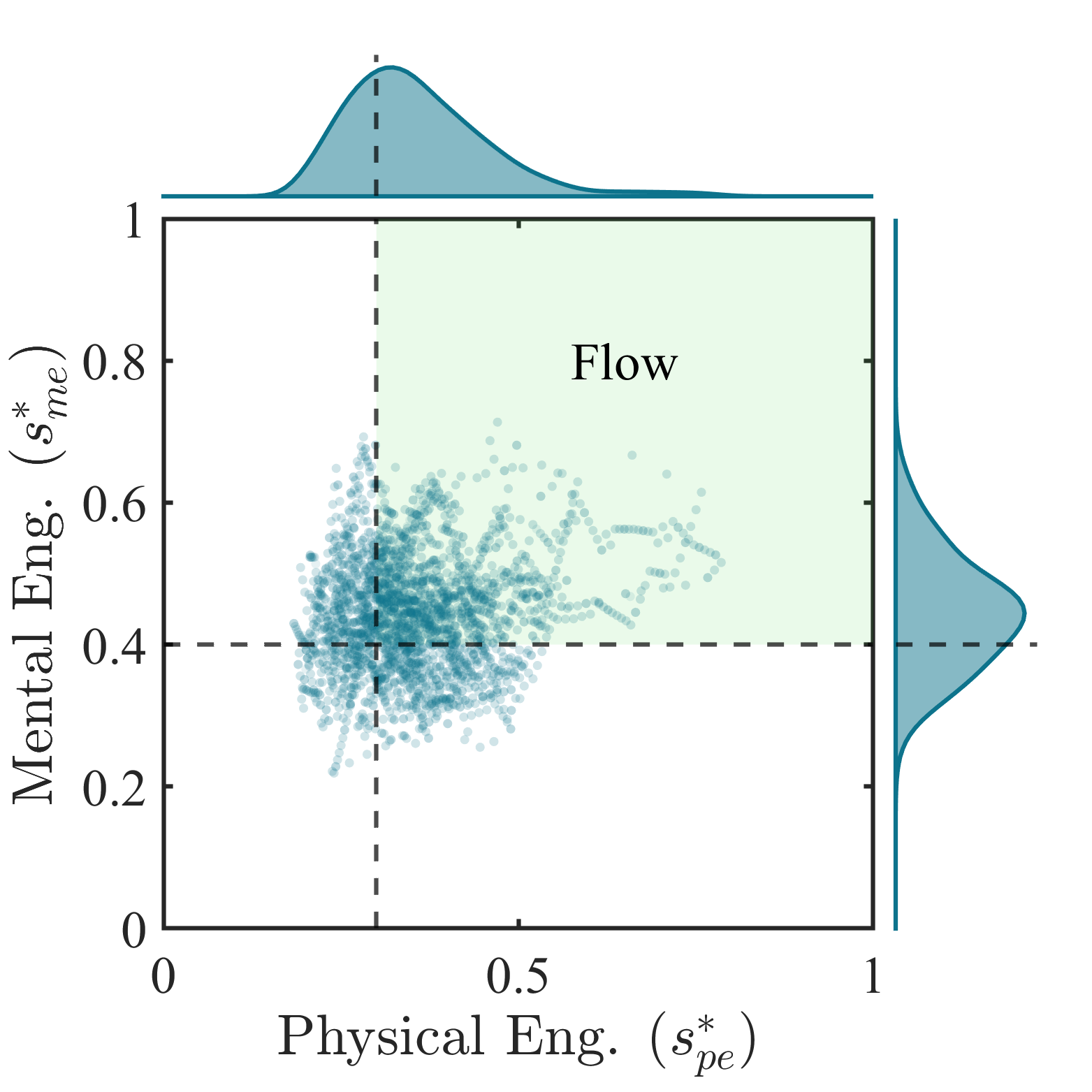}
        \label{fig:kde_fm}
    }
    \caption{Physiological state maintenance and 2D state distribution derived from offline analysis. (a) Grand-averaged mental and physical engagement scores, alongside the Flow State Ratio. (b) 2D Kernel Density Estimation (KDE) map of user states under the baseline AAN mode. (c) 2D KDE map of user states under the proposed Fusion Mode (FM).}
    \label{fig:micro_state}
\end{figure}

Kinematically, while both modes maintained comparable trajectory tracking (Fig.~\ref{fig:macro_performance}(a), (b)), the FM reduced the mean robotic assistive force by $72.4\%$ (Fig.~\ref{fig:macro_performance}(c)). This raw mechanical force reduction provides an objective, independent metric confirming that the FM elicited greater voluntary human effort. Physiologically, 2D Kernel Density Estimation maps show the baseline density drifting into passive lower-left regions (Fig.~\ref{fig:micro_state}(b)). Under the FM, this high-density cluster shifted into the upper-right Flow quadrant (Fig.~\ref{fig:micro_state}(c)). Consequently, the Flow State Ratio, defined as the percentage of total session time where both mental ($s^*_{\text{me}}$) and physical ($s^*_{\text{pe}}$) engagement scores exceeded their respective Gaussian inflection thresholds ($0.4$ and $0.3$), increased nearly tenfold from $4.6\%$ to $42.8\%$.

\begin{figure*}[htbp] 
    \centering
    \captionsetup[subfloat]{captionskip=2pt} 
    
    \subfloat[]{
        \includegraphics[width=0.48\linewidth]{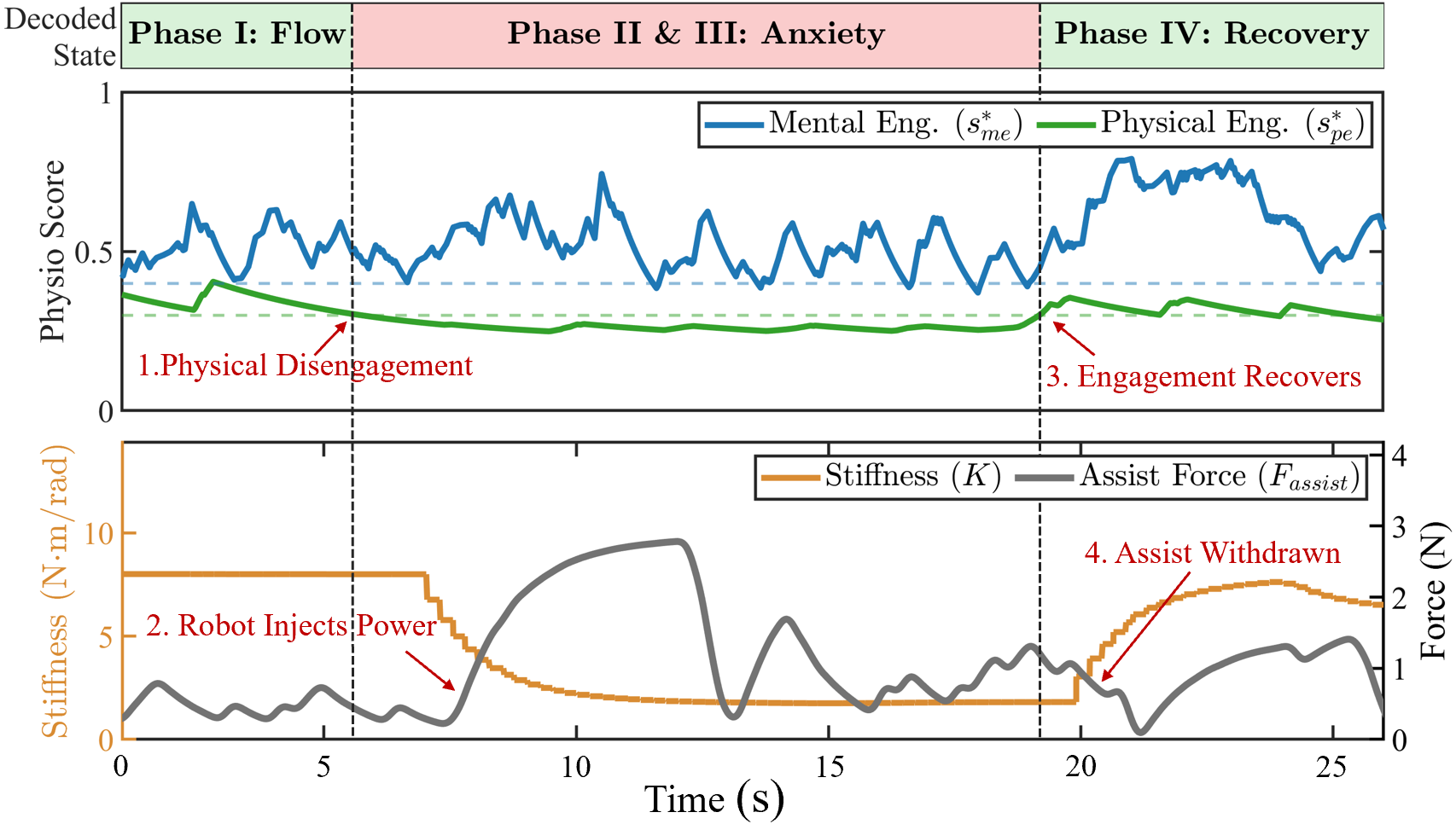}
        \label{fig:example_anxiety}
    }
    \hfill 
    \subfloat[]{
        \includegraphics[width=0.48\linewidth]{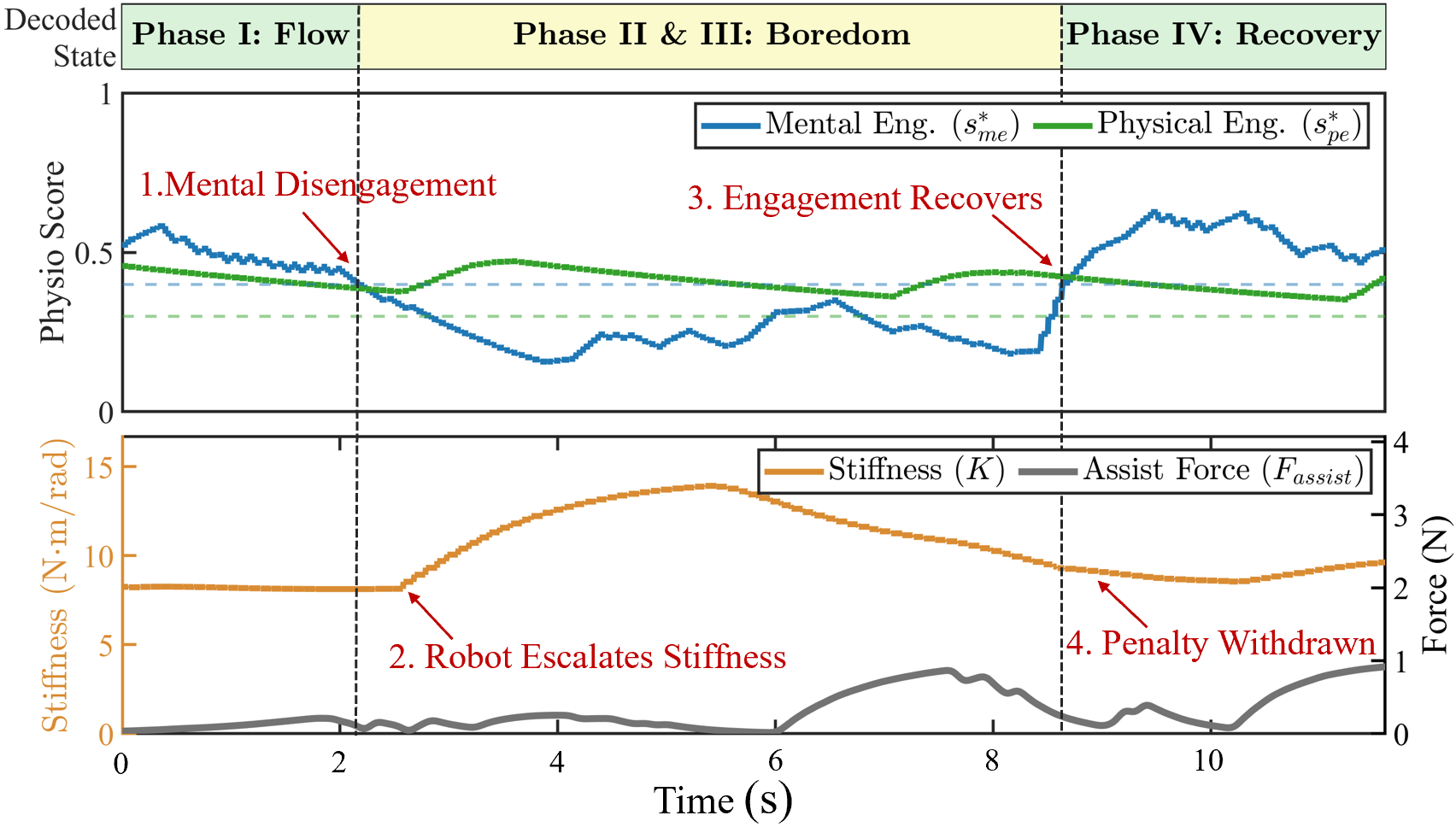}
        \label{fig:example_boredom}
    }
    
    \caption{Microscopic case studies illustrating the closed-loop autonomous recovery mechanisms. \textbf{(a)} Recovery from Anxiety (intent-execution mismatch): the system decreases virtual stiffness ($K$) and elevates feedforward assistance ($F_{\text{assist}}$) in response to acute motor output attenuation. \textbf{(b)} Recovery from Boredom (attentional drift): the system escalates virtual stiffness ($K$) to apply a haptic penalty in response to dropping mental engagement, thereby prompting rapid cortical re-engagement.}
    \label{fig:exp2_example}
\end{figure*}

To detail the closed-loop mechanisms driving these macroscopic gains, two representative microscopic windows were extracted from the first FM trial of Subject 2 (Fig.~\ref{fig:exp2_example}). Time axes are relative to highlight localized transient dynamics. During motor output attenuation (Anxiety, Fig.~\ref{fig:exp2_example}(a)), the system injected feedforward assistance ($F_{\text{assist}}$ spiked to $\sim 3.5\,\text{N}$) to bridge the intent-execution mismatch, allowing the user to gradually regain physical capability over 15--20\,s. Conversely, during visual distraction (Boredom, Fig.~\ref{fig:exp2_example}(b)), the robot escalated virtual stiffness to $\sim 14\,\text{N}\cdot\text{m/rad}$ rather than injecting power. This rigid haptic penalty triggered rapid cortical re-engagement within 5--10\,s.

By decoupling the control outputs, the FM translates multidimensional physiological anomalies into targeted robotic interventions. This dual-intervention mechanism curtails passivity and sustains an environment conducive to neuroplasticity.

\section{Discussion}

\subsection{Resolving Intent-Execution Ambiguity in Multimodal AAN Control}
While recent advancements in robot-assisted rehabilitation have driven significant progress in personalized therapy, sustaining optimal neuroplastic engagement remains a formidable challenge. Contextualizing the proposed architecture within the landscape of contemporary AAN paradigms elucidates the necessity of its underlying design and its fundamental functional departures.

Unimodal sEMG- or kinematic-driven AAN controllers \cite{compare6} adapt support via peripheral effort but remain blind to cognitive states. Failing to distinguish motor fatigue from mental distraction, they inadvertently reward rote repetitions or withhold assistance from exhausted users. Conversely, EEG-driven systems \cite{eeg4flow1} monitor mental engagement but lack peripheral feedback to assess execution capability, unable to discern whether a disengaged user requires feedforward assistance or a haptic penalty. To address this, recent multimodal systems \cite{compare4, compare5} fuse EEG and sEMG, yet typically implement decoupled integrations (e.g., using EEG as a discrete switch while sEMG drives admittance). Such rigid state-machine designs fail to continuously diagnose intent-execution congruence, while their discrete thresholds induce mechanical chattering during physiological fluctuations, disrupting immersion.

The design philosophy of the proposed framework shifts from discrete, symptom-based triggering to a continuous physiological-mechanical closed loop grounded in Flow theory. Rather than layering multimodal signals, it evaluates the cognitive-motor balance via a continuously differentiable ($C^1$) fuzzy manifold. By decoding intent-execution mismatches, it delivers chatter-free haptic interventions, specifically assisting during motor output attenuation (\textit{Anxiety}) and penalizing slacking (\textit{Boredom}). By adapting to the user's state without disrupting immersion, this physiological-mechanical fusion curtails passivity, driving the tenfold increase in neuroplastic engagement ($42.8\%$ vs. $4.6\%$). However, this multimodal approach requires a trade-off in deployment efficiency. Simultaneous EEG and sEMG sensor preparation, combined with the individualized empirical tuning of the fuzzy parameters, introduces higher calibration complexity and computational overhead compared to unimodal systems.

\subsection{Multimodal Integration and Functional Implications}
Integrating hybrid physiological smoothing with the macroscopic Dwell-Time Stabilizer eliminates standard BCI ``chattering,'' yielding safe impedance modulation despite biosignal non-stationarity. The Flow-grounded strategy penalizes antagonist co-activation within the physical metric ($s^*_{\text{pe}}$). Unlike simple force estimation that misinterprets excessive co-contraction as high effort, the proposed metric identifies it as low effective engagement, triggering an \textit{Assist \& Guide} strategy to prevent reinforcing maladaptive patterns. ``Physiological Clamping'' ensures that insufficient challenge triggers \textit{Boredom} haptic penalties, compelling users to train within target physiological zones.

The comparative efficacy results demonstrate the transition from execution-aware to intent-aware training. Common EMG-driven baselines \cite{emg_aan1,emg_aan4} assume muscular output mirrors active participation, rewarding distracted repetitions or wrongly withholding assistance during motor output attenuation. The proposed Fusion Mode resolves this ambiguity. By coupling bidirectional haptic adaptability, specifically penalizing distraction and assisting during motor output attenuation, the system drove a 72.4\% reduction in robotic energy and a nearly tenfold increase in Flow State Ratio, proving that sustaining neuroplasticity demands synchronized central neural drive and peripheral motor effort.

\subsection{Limitations and Future Work}
Validating on healthy young adults limits generalizability to elderly stroke populations, as healthy intent and execution are naturally coupled. The induced mismatches represent controlled approximations. Future longitudinal trials enrolling age-matched stroke survivors will validate parameter generalizability and therapeutic robustness against severe, neurologically derived intent-execution mismatches.

While the Apathy quadrant detects complete disengagement, sustained physical exertion may induce muscle fatigue prior to total mental drop-off. Future deployments will integrate an auxiliary safety layer analyzing real-time sEMG spectral shifts (e.g., median frequency decline) to monitor fatigue directly, preemptively modulating robotic impedance or initiating mandatory rest to ensure patient safety.

\section{Conclusion}
This paper presented a multimodal AAN framework for upper-limb rehabilitation that integrates Flow Theory with adaptive variable impedance control. By fusing real-time EEG and sEMG signals to simultaneously quantify mental and physical engagement, a four-quadrant fuzzy decision engine was developed to mitigate intent-execution mismatch and passive user disengagement. Compared to widely adopted unimodal sEMG-driven baselines, the proposed framework achieved tighter trajectory tracking while reducing robotic energy expenditure by 71.9\%, confirming the elicitation of greater voluntary human effort. Physiologically, it increased the time spent in the optimal neuroplasticity zone (Flow State Ratio) nearly tenfold (from 4.6\% to 42.8\%). Continuous dynamic testing verified the system's stability, demonstrating its ability to safely inject feedforward assistance during motor output attenuation and escalate guiding stiffness during mental distraction. This decoupled, sensorless architecture provides a safe, intent-responsive, and intensive human-robot interaction.

\vspace{0.1cm}
\nopagebreak
\interlinepenalty=10000

\bibliographystyle{ieeetr}
\bibliography{sample}

\end{document}